\documentclass[a4paper,fleqn]{cas-dc}

\usepackage[authoryear]{natbib}
\usepackage{url}

\usepackage{graphicx}
\usepackage{subfigure}
\usepackage{amsmath,amssymb,amsfonts}
\usepackage{algorithmic}
\usepackage{color}
\usepackage{bm}
\usepackage{algorithm}

\def\tsc#1{\csdef{#1}{\textsc{\lowercase{#1}}\xspace}}
\tsc{WGM}
\tsc{QE}
\tsc{EP}
\tsc{PMS}
\tsc{BEC}
\tsc{DE}

\begin{document}
\let\WriteBookmarks\relax
\def\floatpagepagefraction{1}
\def\textpagefraction{.001}
\shorttitle{Compressing 3DCNNs Based on Tensor Train Decomposition}
\shortauthors{Dingheng Wang et~al.}

\title [mode = title]{Compressing 3DCNNs Based on Tensor Train Decomposition}                      

\author[1]{Dingheng Wang}
\fnmark[1]
\address[1]{School of Electronic and Information Engineering, Xi'an Jiaotong University, Xi'an 710049, China}

\author[1]{Guangshe Zhao}
\fnmark[2]

\author[2]{Guoqi Li}
\cormark[1]
\fnmark[3]
\address[2]{Department of Precision Instrumentation, Center for Brain Inspired Computing Research and  Beijing Innovation Center for Future Chip, Tsinghua University, Beijing 100084, China}

\author[3]{Lei Deng}
\fnmark[4]
\address[3]{University of California, Santa Barbara, CA93106, USA}

\author[4]{Yang Wu}
\fnmark[5]
\address[4]{Institute for Research Initiatives, Nara Institute of Science and Technology, Takayama-cho, Ikoma, Nara, Japan}

\cortext[cor1]{Corresponding author}
\fntext[fn1]{wangdai11@stu.xjtu.edu.cn}
\fntext[fn2]{zhaogs@mail.xjtu.edu.cn}
\fntext[fn3]{liguoqi@mail.tsinghua.edu.cn}
\fntext[fn4]{leideng@ucsb.edu}
\fntext[fn5]{yangwu@rsc.naist.jp}

\begin{abstract}
Three dimensional convolutional neural networks (3DCNNs) have been applied in many tasks, e.g., video and 3D point cloud recognition. However, due to the higher dimension of convolutional kernels, the space complexity of 3DCNNs is generally larger than that of traditional  two dimensional convolutional neural networks (2DCNNs). To miniaturize 3DCNNs for the deployment in confining environments such as embedded devices, neural network compression is a promising approach. In this work, we adopt the tensor train (TT) decomposition, a straightforward and simple \emph{in situ} training compression method, to shrink the 3DCNN models. Through proposing tensorizing 3D convolutional kernels in TT format, we investigate how to select appropriate TT ranks for achieving higher compression ratio. We have also discussed the redundancy of 3D convolutional kernels for compression, core significance and future directions of this work, as well as the theoretical computation complexity versus practical executing time of convolution in TT. In the light of multiple contrast experiments based on VIVA challenge, UCF11, and UCF101 datasets, we conclude that TT decomposition can compress 3DCNNs by around one hundred times without significant accuracy loss, which will enable its applications in extensive real world scenarios. 
\end{abstract}


\maketitle

\section{Introduction}\label{sec:Intro}

In the past few years, deep neural networks (DNNs) \citep{LeCun_2015_DNN} have achieved great success in machine learning, especially the convolutional neural networks (CNNs) with some representative instances such as AlexNet \citep{Krizhevsky_2012_AlexNet}, VGG-Net\citep{Simonyan_2015_VGG}, Goo\-gleNet \citep{Szegedy_2015_GoogleNet}, ResNet \citep{He_2016_ResNet}, Dense\-Net \citep{Huang_2017_DenseNet}, etc. Nowadays, three dimensional convolutional neural networks (3DCNNs) \citep{Ji_Shuiwang_2013_3DCNN,Tran_2015_3DCNN} have been applied in many tasks of recognition of spatio-temporal data from videos \citep{Zhang_2017_3DCNNLSTM,Zhu_2017_3DCNN_LSTM,Molchanov_2015_3DCNN_1,Molchanov_2015_3DCNN_2,Molchanov_2016_3DCNN_3,Camgoz_2016_3DCNN,Varol_2018_LongTerm3DCNN,Hara_2018_Res3DCNN}, pure 3D data from depth cameras \citep{Ge_2017_PointsCloud3DCNN}, and stacking utterances from speech data \citep{Torfi_2018_Speak3DCNN}. However, these high-dimensional 3DCNN architectures, e.g., long-term temporal convolutions with large sized 3D convolutional kernels \citep{Varol_2018_LongTerm3DCNN}, make the situation of inflated sizes of DNNs \citep{Cheng_2018_CompressSurvey} more serious. Even worse, to the best of our knowledge, there are few practices to compress 3DCNNs to satisfy miniaturization requirements in confining environments such as embedded devices.

Fortunately, there are researches on the compression of other neural networks \citep{Cheng_2018_CompressSurvey}, which provide opportunity to compress 3DCNNs, e.g., compact architecture, weight sharing or quantization, sparsifying or pruning, knowledge distillation, and matrix/tensor decomposition or low-rank factorization. Among these methods, compact architecture can just obtain limited compression ratio and elaborate design is necessary, quantization maps the weights from floating numbers to integers to accelerate computation so that its compression ratio is not very high, pruning usually needs to pre-train corresponding uncompressed models and the data structure of pruned weight appears to be intricate thus extra marked data may be indispensable, and distillation is generally inefficient in training since two networks should be dealt with. By contrast, decomposition method may afford us the so-called \emph{in situ} training \citep{Alibart_2013_ExInSitu} which can directly get a trained model from scratch with sufficient compression performance because of its inherent theory of linear algebra. Therefore, in this work, we focus on compressing 3DCNNs by applying decomposition methods.

In the aspect of decomposition methods, singular value decomposition (SVD) is the most widely employed matrix decomposition method for the compression of DNNs. For instance, \citet{Zhang_2015_SVD,Zhang_2016_SVD} split one convolutional kernel into two sub kernels, and \citet{Shim_2017_SVD} compress the last softmax layer for neural networks with large vocabulary. However, it may be not enough to completely eliminate the inherent redundancy in DNNs \citep{Denil_2013_Redundancy} from the point view  of tensor. Hence, a higher compression ratio could be approached by reshaping the weight matrices to tensors, termed as \emph{tensorizing} \citep{Novikov_2015_TT}. Nevertheless, traditional tensor decomposition methods, such as CP \citep{Caroll_1970_CP} and Tucker \citep{Tucker_1966_Tucker}, are inevitable to fall in the curse of dimensionality because their kernel tensors still give an exponential contribution to the space complexity \citep{Cichocki_2015_TensorApp}.

Tensor network decomposition methods \citep{Cichocki_2018_TensorNetworks}, including hierarchical Tucker \citep{Hackbusch_2009_InventHT,Grasedyck_2010_InventHT}, tensor train (TT) \citep{Oseledets_2011_InventTT}, and tensor chain \citep{Khoromskij_2011_InventTC,Zhao_2018_TR}, can completely avoid the curse of dimensionality   by representing a tensor as linked tiny factor tensors with restricted orders. Thereinto, TT decomposition is the most concise format so that many compression applications are based on it. \citet{Novikov_2015_TT} first utilize TT decomposition to compress the weight matrices in fully connected (FC) layers. Since then, \citet{Huang_2018_TTCNN}, \citet{Su_2018_TTCNN} and \citet{Huang_2019_TTCNN} extend this idea to the applications based on CNNs with TT decomposed FC layers. Only \citet{Garipov_2016_TTCNN} apply the TT format to convolutional layers by first viewing the kernel as a 4th-order tensor, then reshaping the tensor to a matrix, and finally matching the matrix to the \(d\)th-order tensorizing TT approach \citep{Novikov_2015_TT}. In the domain of recurrent neural networks (RNNs), \citet{Tjandra_2017_TTRNN1,Tjandra_2018_TTRNN2} utilize TT format to compress all matrices within different kinds of gated structures, and further, \citet{Yang_2017_TTRNN} test the performance of TT-RNNs and achieve extremely high compression ratio with miraculous accuracy improvement rather than loss based on larger models and datasets.

From the above recent practices on TT-based compression, we observe that: 1) all but \citet{Garipov_2016_TTCNN}'s method are based on tensorizing TT approach for weight matrices including FC layers and RNN gated units; 2) all but \citet{Yang_2017_TTRNN}'s work have more or less accuracy losses; 3) although it is important to make TT format to be low-rank \citep{Lee_2014_TTRanks,Bengua_2017_TTRanks}, how to select suitable TT ranks with given tensor shape especially for training DNNs has not been addressed yet. Based on these observations and facing the specific 3D convolutional kernels in 3DCNNs, in this work, we will first study a tensorizing method to compress a 5th-order 3D convolutional kernel tensor into the TT format as a \(d\)th-order tensor according to fundamental methods reported by \citet{Novikov_2015_TT} and \citet{Garipov_2016_TTCNN}. Secondly, we will provide a general rule to decide the values of TT ranks for a specific tensor with given shape. Thirdly, inspired by \citet{Yang_2017_TTRNN}, multiple experiments on VIVA challenge \citep{Ohn-Bar_2014_VIVA}, UCF11 \citep{Liu_2011_UCF11} and UCF101 \citep{Soomro_2012_UCF101} datasets will be conducted to give empirical proof that accuracy loss can be avoided in TT compressed 3DCNNs with around one hundred times compression ratio if their original uncompressed ones have comparatively higher level of redundancy, i.e., larger scale of networks. Last but not least, some other characteristics of TT will also be discussed to draw forth the core significance and future direction of TT CNNs, e.g., regularization derived from fewer parameters of TT, and theoretical computation complexity versus practical executing time of convolution in TT.

We list the main contributions of this work as follows.
\begin{itemize}
\item To the best of our knowledge, we are the first to utilize the TT format to compress convolutional kernels in 3DCNNs. This method can provide an direct \emph{in situ} training approach without pre-training or elaborate design to compress large-scale 3DCNNs for application scenarios with limited storage space.

\item We establish a general principle to select TT ranks for a size-fixed tensor based on two bases. One is the theoretical analysis to explain the source of TT ranks which come from hierarchical Tucker decomposition, and the other is the experimental verification.

\item We empirically demonstrate that the accuracy loss in compression can be avoided in 3DCNNs with high redundancy by combining the inherent regularity of TT decomposition, and further a very high compression ratio (about one hundred times) can be obtained.
\end{itemize}

The rest of the paper is organized as follows. Section \ref{sec:Method} first introduces fundamental knowledge of the TT format including tensorizing for matrices, then proposes the tensorizing for 3D convolutional kernels and discusses the selection of TT ranks. Section \ref{sec:Exp} presents the elaborate contrast experiments based on VIVA challenge, UCF11 and UCF101 datasets to verify that the accuracy loss can be avoided when compressing a redundant 3DCNN model based on TT decomposition. Section \ref{sec:Dis} further discusses some experimental phenomena and possible internal mechanisms of TT 3DCNNs. Section \ref{sec:Con} concludes this work and mentions the future direction.

\section{Tensor Train Decomposition for 3DCNNs}\label{sec:Method}

In this section, we first introduce basic knowledge of TT format. Then we propose the tensorizing method for compressing 3D convolutional kernels. Finally we investigate the principle  regarding  how to select TT ranks. For convenience, we will use the bold lower case letter as the vector symbol (e.g. \( \bm{a} \)), the bold upper case letter as the matrix symbol (e.g. \( \bm{A} \)), the calligraphic bold upper case letter as the tensor notation (e.g. \( \bm{ \mathcal{A} } \)), and the corresponding ordinary letters (e.g. \(a\), \(A\), and \(\mathcal{A}\)) to denote their respective elements.

\subsection{Tensor Train Format}

\subsubsection{Basic TT Format}\quad

According to \citet{Oseledets_2011_InventTT}, the basic TT format of a \(d\)th-order tensor \(\bm{\mathcal{A}} \in \mathbb{R} ^{n_{1} \times n_{2} \times \cdots \times n_{d}}\) can be represented in the measure of entry as
\begin{equation}\label{Eq_TTBaseElement}
\mathcal{A}(j_{1},j_{2},\cdots,j_{d}) = \bm{G}_{1}[j_{1}] \bm{G}_{2}[j_{2}] \cdots \bm{G}_{d}[j_{d}]
\end{equation}
where \( j_{k} \in \{1,2,\cdots,n_{k}\} \) (\( k \in \{1,2,\cdots,d\} \)) is the \( k \)th index of the entry in tensor \( \bm{\mathcal{A}} \), serial products on the right side of the equation are core matrices to calculate the entry, each matrix \( \bm{G}_{k}[j_{k}] \) has the shape of \( r_{k-1} \times r_{k} \) and \( r_{0}=r_{d}=1 \). There are totally \( d+1 \) values of \( r_{k} \) which are collectively called TT ranks.  Additionally, all \( \bm{G}_{k}[j_{k}] \) corresponding to the same mode \(n_{k}\) can be stacked into a \( 3 \)rd-order core tensor \( \bm{\mathcal{G}}_{k} \in \mathbb{R} ^{r_{k-1} \times n_{k} \times r_{k}} \). Therefore, the TT format of \(\bm{\mathcal{A}}\) can also be represented as \citep{Lee_2016_HTTT}
\begin{equation}\label{Eq_TTBaseTensor}
\bm{\mathcal{A}} = \bm{\mathcal{G}}_{1} \times ^{1} \bm{\mathcal{G}}_{2} \times ^{1} \cdots \times ^{1} \bm{\mathcal{G}}_{d}
\end{equation}
where \( \times ^{1} \) is called mode-\( (N,1) \) contracted product, which means just one pair equal modes in any \( N \)th-order tensor \( \bm{\mathcal{X}} \) and \( M \)th-order tensor \( \bm{\mathcal{Y}} \) will be contracted to produce a new \( (N+M-2) \)th-order tensor \(\bm{\mathcal{X}} \times ^{1} \bm{\mathcal{Y}}\).

Suppose that the maximum value of all modes is \(n\), and the maximal rank is \(r\). It is easy to work out that the space complexity of tensor \(\bm{\mathcal{A}}\) can be reduced from \(\mathcal{O}(n^{d})\) to \(\mathcal{O}(dnr^2)\). Obviously, the compression ratio grows exponentially as the value of order \(d\) increases linearly. This means that the more complex a data structure is, the higher compression ratio we can obtain.

\subsubsection{Tensorizing and TT for Matrices}\quad

\begin{figure*}
\centering
\subfigure[Tensorizing for weight matrix in TT format]{\includegraphics[width=0.9\textwidth]{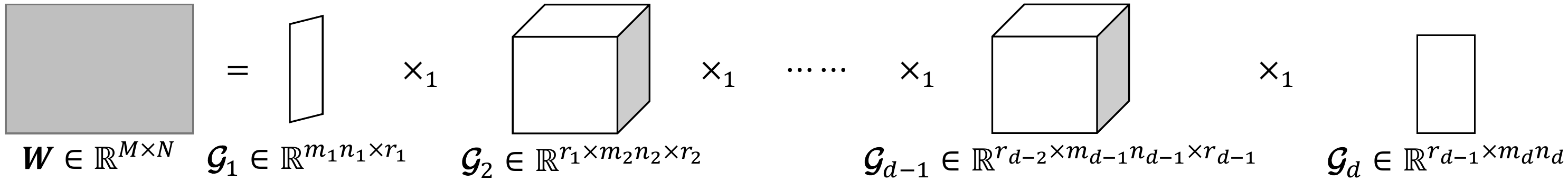}}
\subfigure[Tensorizing for convolutional kernel in TT format]{\includegraphics[width=0.9\textwidth]{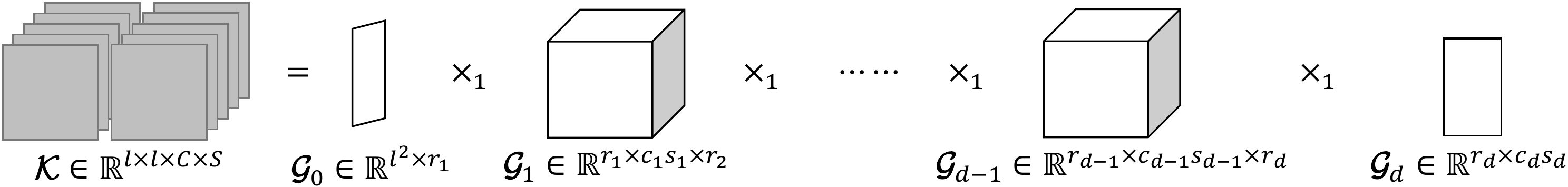}}
\subfigure[Tensorizing for 3D convolutional kernel in TT format]{\includegraphics[width=0.9\textwidth]{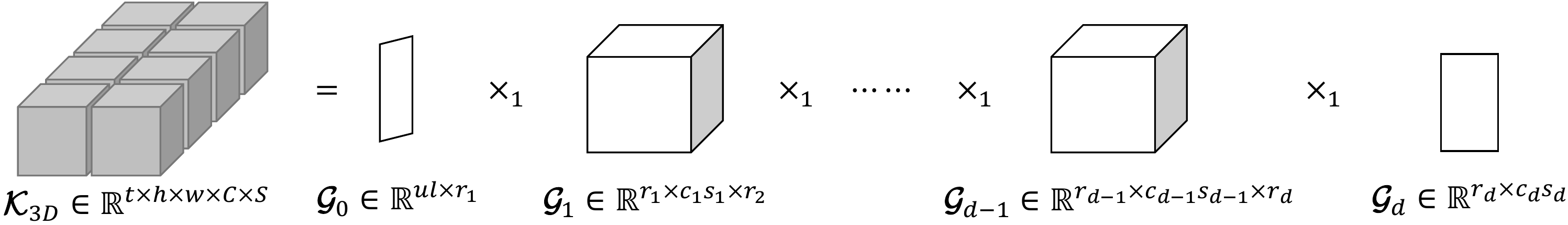}}
\caption{\textbf{TT decomposition structures of weight matrix, convolutional kernel, and 3D convolutional kernel,} where \(M = \prod_{i=1}^d m_i\), \(N = \prod_{i=1}^d n_i\), \(C = \prod_{i=1}^d c_i\), \(S = \prod_{i=1}^d s_i\) and \(t\times h \times w=u \times l\).}
\label{Fig_tt_structure}
\end{figure*}

It is meaningless to directly use Equation (\ref{Eq_TTBaseTensor}) to decompose a matrix as a 2nd-order tensor, because such naive approach will make TT decomposition degenerating as normal low-rank matrix decomposition. From analysis of space complexity above, significant compression ratio can be obtained if the original matrix is reshaped to set the value of order higher. Such idea is the so-called \emph{tensorizing} \citep{Novikov_2015_TT}, which makes it possible to utilize TT for matrices not only in the field of DNNs.

In detail, consider a large matrix \( \bm{W} \in \mathbb{R} ^{M \times N} \) and each value of its modes can be factorized into \( d \) integers like \(M=\prod_{i=1}^{d} m_{i}\) and \(N=\prod_{i=1}^{d} n_{i}\) (\( i \in \{1,2,\cdots,d\} \)). Then a \( d \)th-order tensor \( \bm{ \mathcal{W} } \in \mathbb{R} ^{m_{1}n_{1} \times m_{2}n_{2} \times \cdots \times m_{d}n_{d}} \) can be constructed by two bijections which can map the original matrix mode \( M \) or \( N \) to \( d \) tensor modes \( m_{i} \) or \( n_{i} \) separately. The corresponding relationship between the original matrix and the reshaped tensor can be represented as
\begin{equation*}
\begin{aligned}
W(\zeta,\xi) = \mathcal{W}((\nu_{1}(\zeta),\mu_{1}(\xi)),(\nu_{2}(\zeta),\mu_{2}(\xi)),\cdots,(\nu_{d}(\zeta),\mu_{d}(\xi)))
\end{aligned}
\end{equation*}
where \( \zeta \in \{1, 2, \cdots, M\} \), \( \xi \in \{1, 2, \cdots, N\} \), and
\begin{equation*}
\begin{aligned}
\bm{\nu}(\zeta) &= (\nu_{1}(\zeta), \nu_{2}(\zeta), \cdots, \nu_{d}(\zeta)) \\
\bm{\mu}(\xi) &= (\mu_{1}(\xi), \mu_{2}(\xi), \cdots, \mu_{d}(\xi))
\end{aligned}
\end{equation*}
are the two bijections.

After tensorizing, one can use Equation (\ref{Eq_TTBaseElement}) to rewrite the tensor \(\bm{\mathcal{W}}\) into its TT format as
\begin{equation}\label{Eq_TTMatrix}
\begin{aligned}
&\mathcal{W}((\nu_{1}(\zeta),\mu_{1}(\xi)),\cdots,(\nu_{d}(\zeta),\mu_{d}(\xi))) = \\
&\bm{G}_1[(\nu_{1}(\zeta),\mu_{1}(\xi))]\cdots\bm{G}_d[(\nu_{d}(\zeta),\mu_{d}(\xi))].
\end{aligned}
\end{equation}
Relevant space complexity can be reduced from \(\mathcal{O}(MN)\) to \(\mathcal{O}(dmnr^2)\) where \(m\) and \(n\) are the maximal \(m_i\) and \(n_i\) (\(i \in \{1,2,\cdots,d\}\)), respectively. The visualized structure of TT for matrix   \( \bm{W} \in \mathbb{R} ^{M \times N} \)  is illustrated in Figure \ref{Fig_tt_structure}(a).

\subsection{TT for 3D Convolutional Kernels}

\subsubsection{From 2D to 3D}\quad

\begin{algorithm}[htb] 
\caption{3D convolution with kernel in TT format.}
\label{Alg_TT_3DCNN} 
\begin{algorithmic}[1]
\REQUIRE ~~\\
Input data, \(\bm{\mathcal{I}} \in \mathbb{R} ^{T \times H \times W \times C}\);\\
Output channel, \(S\);\\
Modes of input channels, \(\bm{c}=[c_1,c_2,\cdots,c_d]\);\\
Modes of output channels, \(\bm{s}=[s_1,s_2,\cdots,s_d]\);\\
Ranks, the same count as \(\bm{c}\) and \(\bm{s}\), \(\bm{r}=[r_1,r_2,\cdots,r_d]\);\\
Convolutional filter shape, \(\bm{k}_{filter}=[t,h,w]\).
\ENSURE ~~\\
Output data by convolution with padding so that the shape of single channel (\(T \times H \times W\)) is invariant, \(\bm{\mathcal{O}} \in \mathbb{R} ^{T \times H \times W \times S}\).
\STATE \(p \longleftarrow thw\);
\STATE \(q \longleftarrow \sqrt{p}\);
\STATE Gain the nearest lower integer of \(q\), \(l \longleftarrow\) lower(\(q\));
\STATE Gain the nearest upper integer of \(q\), \(u \longleftarrow\) upper(\(q\));
\WHILE{True}
\IF{\(p \bmod u = 0\)}
\STATE \(l \longleftarrow p/u\);
\STATE Break;
\ELSIF{\(p \bmod l = 0\)}
\STATE \(u \longleftarrow p/l\);
\STATE Break;
\ELSE
\STATE \(l \longleftarrow l-1\);
\STATE \(u \longrightarrow u+1\);
\ENDIF
\ENDWHILE
\STATE Define \(\overline{\bm{\mathcal{G}}}_{0} \in \mathbb{R} ^{ul \times r_1}\) in disk;
\FOR{\(i = 1 \longrightarrow d-1\)}
\STATE Define \(\bm{\mathcal{G}}_{i} \in \mathbb{R} ^{r_{i} \times c_{i}s_{i} \times r_{i+1}}\) in disk;
\ENDFOR
\STATE Define \(\bm{\mathcal{G}}_{d} \in \mathbb{R} ^{r_{d} \times c_{d}s_{d}}\) in disk;
\STATE \(\overline{\bm{\mathcal{K}}}_{3D} \longleftarrow \overline{\bm{\mathcal{G}}}_{0} \times^{1} \bm{\mathcal{G}}_{1} \times^{1} \bm{\mathcal{G}}_{2} \times^{1} \cdots \times^{1} \bm{\mathcal{G}}_{d}\);
\STATE Reshape as \citep{Garipov_2016_TTCNN}, \(\widetilde{\bm{\mathcal{K}}} _{3D} \longleftarrow \overline{\bm{\mathcal{K}}}_{3D}\);
\STATE Reshape as Equation (\ref{Eq_Map_3DCNN}), \(\bm{\mathcal{K}} _{3D} \longleftarrow \widetilde{\bm{\mathcal{K}}}_{3D}\);
\STATE \(\bm{\mathcal{O}} \longleftarrow \bm{\mathcal{I}} * \bm{\mathcal{K}} _{3D}\);
\RETURN \(\bm{\mathcal{O}}\).
\end{algorithmic}
\end{algorithm}

A normal 2D convolutional kernel could always be regarded as a 4th-order tensor \(\bm{\mathcal{K}} \in \mathbb{R} ^{l \times l \times C \times S}\), which can be reshaped to a \((d+1)\)th-order tensor \(\overline{\bm{\mathcal{K}}} \in \mathbb{R} ^{l^2 \times c_{1}s_{1} \times c_{2}s_{2}}\) \(^{\times \cdots \times c_{d}s_{d}}\) by referring to tensorizing for matrices, where \(l\) means the edge length of convolutional filter, \(C = \prod_{i=1}^d c_i\) and \(S = \prod_{i=1}^d s_i\) denote the input and the output channels respectively \citep{Garipov_2016_TTCNN}. For TT format, such reshaping approach is more efficient than naively utilizing Equation (\ref{Eq_TTBaseElement}), because the value of \(C\) or \(S\) is usually much larger than \(l\), and a tensor with more balanced shape can usually get less errors \citep{Novikov_2015_TT}. Based on Equation (\ref{Eq_TTMatrix}), \citet{Garipov_2016_TTCNN} give the TT format for convolutional kernels
\begin{equation}\label{Eq_TTCNN}
\begin{aligned}
&\overline{\mathcal{K}}((l_a^{'},l_b^{'}),(c_1^{'},s_1^{'})\cdots,(c_d^{'},s_d^{'})) = \\
&\overline{\bm{G}}_0[(l_a^{'},l_b^{'})]\bm{G}_1[(c_1^{'},s_1^{'})]\cdots\bm{G}_d[(c_d^{'},s_d^{'})]
\end{aligned}
\end{equation}
where \(l_a^{'}\) or \(l_b^{'} \in \{1,2,\cdots,l\}\), \(c_i^{'} \in \{{1,2,\cdots,c_i}\}\), and \(s_i^{'} \in \{{1,2,\cdots,s_i}\}\) \((i \in \{1,2,\cdots,d\})\). The visualized structure of TT for convolutional kernel \(\bm{\mathcal{K}} \in \mathbb{R} ^{l \times l \times C \times S}\) is illustrated in Figure \ref{Fig_tt_structure}(b).

Inspired by \citet{Garipov_2016_TTCNN}, we intend to propose a similar method to reconstruct a 3D convolutional kernel to a \((d+1)\)th-order tensor with relatively balanced shape, and then utilize the TT format on this tensor. However, a 3D convolutional kernel \(\bm{\mathcal{K}} _{3D} \in \mathbb{R} ^{t \times h \times w \times C \times S}\) is a 5th-order tensor which has a convolutional filter with 3 sizes (\(t \times h \times w\)) rather than regular hexahedron in most cases. Thus, in order to utilize Equation (\ref{Eq_TTCNN}), we should make a mapping to transfer the entry from \(\bm{\mathcal{K}} _{3D}\) to a new 4th-order tensor \(\widetilde{\bm{\mathcal{K}}} _{3D} \in \mathbb{R} ^{u \times l \times C \times S}\) with the constraint \(t \times h \times  w = u \times  l\).

First, let us ignore \(C\) and \(S\), stretch \(\bm{\mathcal{K}} _{3D}\) to a 3rd-order tensor \(\widehat{\bm{\mathcal{K}}} _{3D} \in \mathbb{R} ^{p \times C \times S}\) with the constraint \(thw = p\). Then, suppose the value of each index begins at 0, and we have
\begin{equation*}
\mathcal{K}_{3D}(t^{'},h^{'},w^{'},C^{'},S^{'})=\widehat{\mathcal{K}}_{3D}(p^{'},C^{'},S^{'})
\end{equation*}
where \(t^{'}\), \(h^{'}\), \(w^{'}\), \(C^{'}\), \(S^{'}\), \(p^{'}\) are indices of the modes \(t\),\(h\), \(w\), \(C\), \(S\), \(p\), respectively, with \(w^{'}ht + h^{'}t + t^{'} = p^{'}\). Second, fold \(\widehat{\bm{\mathcal{K}}}_{3D}\) to a 4th-order tensor \(\widetilde{\bm{\mathcal{K}}} _{3D} \in \mathbb{R} ^{u \times l \times C \times S}\) with the constraint \(p=ul\), we have
\begin{equation*}
\widehat{\mathcal{K}}_{3D}(p^{'},C^{'},S^{'})=\widetilde{\mathcal{K}}_{3D}(u^{'},l^{'},C^{'},S^{'})
\end{equation*}
where \(u^{'},l^{'}\) are indices of the modes \(u,l\), with \(p^{'}=l^{'}u+u^{'}\). Combining the above two steps, the mapping from \(\bm{\mathcal{K}} _{3D}\) to \(\widetilde{\bm{\mathcal{K}}} _{3D}\) should include 3 bijections: \(\bm{e}(\zeta) = (e_1(\zeta),e_2(\zeta))\), \(\bm{f}(\eta) = (f_1(\eta),f_2(\eta))\), and \(\bm{g}(\xi) = (g_1(\xi),g_2(\xi))\), which let
\begin{equation}\label{Eq_Map_3DCNN}
\begin{aligned}
&\mathcal{K}_{3D}(\zeta,\eta,\xi,C^{'},S^{'})=\\
&\widetilde{\mathcal{K}}_{3D}((e_1(\zeta),f_1(\eta),g_1(\xi)),(e_2(\zeta),f_2(\eta),g_2(\xi)),C^{'},S^{'})
\end{aligned} 
\end{equation}
and
\begin{equation*}
\xi ht + \eta t + \zeta = (e_2(\zeta),f_2(\eta),g_2(\xi))u + (e_1(\zeta),f_1(\eta),g_1(\xi)).
\end{equation*}

Finally, as the same as Equation (\ref{Eq_TTCNN}), by further reshaping \(\widetilde{\bm{\mathcal{K}}}_{3D}\) to \(\overline{\bm{\mathcal{K}}}_{3D} \in \mathbb{R} ^{ul \times c_{1}s_{1} \times c_{2}s_{2} \times \cdots \times c_{d}s_{d}}\), we can get the TT format for 3D convolutional kernels like
\begin{equation}\label{Eq_TT_3DCNN}
\begin{aligned}
&\overline{\mathcal{K}}_{3D}((u^{'},l^{'}),(c_1^{'},s_1^{'})\cdots,(c_d^{'},s_d^{'})) = \\
&\overline{\bm{G}}_0[(u^{'},l^{'})]\bm{G}_1[(c_1^{'},s_1^{'})]\cdots\bm{G}_d[(c_d^{'},s_d^{'})].
\end{aligned}
\end{equation}

The visualized structure of TT for 3D convolutional kernel   \(\bm{\mathcal{K}} _{3D} \in \mathbb{R} ^{t \times h \times w \times C \times S}\)  is illustrated in Figure \ref{Fig_tt_structure}(c). Moreover, for easily programming and directly using convolutional operation \(*\), the proposed \textbf{Algorithm} \ref{Alg_TT_3DCNN} shows how to design the structure of a 3D convolutional kernel and compute the convolutional output with input data. Note that we calculate the values of \(u\) and \(l\) as close as possible to ensure the data ranges of \(u^{'}\) and \(l^{'}\) are sufficient.

\subsubsection{Compression Ratio and TT Rank}\quad

The corresponding compression ratio of Equation (\ref{Eq_TT_3DCNN}) can be calculated as
\begin{equation*}
R = \frac{ul\prod_{i=1}^d c_{i}s_{i}}{ulr_{1} + \sum_{i=1}^{d-1} c_{i}s_{i}r_{i}r_{i+1} + c_{d}s_{d}r_{d}}
\end{equation*}
where \(r_{i}\) (\(i \in {1,\cdots,d}\)) denotes the TT ranks \(\bm{r}=[r_1,\cdots,r_d]\). Note that \(r_0=r_{d+1}=1\) because \(\overline{\bm{\mathcal{K}}}_{3D}\) is a \((d+1)\)th-order tensor. It is easy to see that the compression ratio depends on the TT ranks significantly. Let \(r^{(u)}\) and \(r^{(l)}\) be the maximum and minimum \(r_{i}\) in \(\bm{r}\), we obtain that 
\begin{equation}\label{Eq_CompressRatio_3DCNN}
\begin{aligned}
&B(r) = \frac{ul\prod_{i=1}^d c_{i}s_{i}}{r(ul + r\sum_{i=1}^{d-1} c_{i}s_{i}  + c_{d}s_{d})} \\
&B(r^{(l)}) \geq R \geq B(r^{(u)}).
\end{aligned}
\end{equation}
This means that TT ranks \(\bm{r}\) influences the range of compression ratio under a certain tensorizing format. As a rule of thumb, higher rank may signify lower error while lower rank may cause more information loss. Detailedly, if we select high rank, the practical compression ratio \(R\) will be very close to the lower bound \(B(r^{(u)})\) which can certainly hinder fulfilling the meaning of compression. On the contrary, if we choose small rank to pursue extremely high compression ratio \(R\) which follows the upper bound \(B(r^{(l)})\) closely, considerable accuracy loss may occur. That is, the trade-off between compression ratio \(R\) and accuracy is critically based on \(\bm{r}\). Thus, how to decide the values of TT ranks has significant correlation to low loss compressing 3DCNNs, and becomes the remaining core topic of this work.

\subsection{The Selection of TT ranks}

\subsubsection{Theoretical Foundation}\quad

As mentioned in Section \ref{sec:Intro}, there is still a lack of verified principles for the selection of TT ranks to represent a tensor with given shape, although \citet{Novikov_2015_TT} have summarized some phenomena when TT ranks growing. In fact, it is widely accepted that the TT format is a special form of the hierarchical Tucker format \citep{Grasedyck_2010_InventHT,Lee_2016_HTTT,Grasedyck_2011_HTTT,Khrulkov_2018_TTRNN}, so it is possible to find the theoretical foundation of TT ranks from researching the details of hierarchical Tucker.

In order to explain the hierarchical Tucker format, we first introduce the concept of \(t\)-modes matricization \citep{Grasedyck_2010_InventHT}. Consider a \( d \)th-order tensor \( \bm{ \mathcal{A} } \in \mathbb{R} ^{n_{1} \times n_{2} \times \cdots \times n_{d}} \) with the set of indices of modes \( u=\{1,2,\cdots,d\} \) which can be split into two subsets \( t=\{t_{1},\cdots,t_{k}\} \) and \( s=\{s_{1},\cdots,s_{d-k}\} \) (\(u=t \cup s\) ). If the set \( t \)  can also be split into \( t=t_{l} \cup t_{r} \), we can conclude \citep{Kressner_2011_HT}
\begin{equation*}
\mathrm{span}(\bm{A} ^{(t)}) \subset \mathrm{span}(\bm{A} ^{(t_{l})} \otimes \bm{A} ^{(t_{r})})
\end{equation*}
where \( \bm{A} ^{(t)} \in \mathbb{R} ^{n_{t_{1}}n_{t_{2}} \cdots n_{t_{k}} \times n_{s_{1}}n_{s_{2}} \cdots n_{s_{d-k}}} \) and \( \otimes \) means the Kronecker product. Similarly, modes of \( \bm{A} ^{(t_{l})} \) are serial products of elements in \( t_{l} \) and its complementary set \( u \setminus t_{l} \), separately, and so does \( \bm{A} ^{(t_{r})} \). Such form like \( \bm{A} ^{(t)} \) is called \( t \)-modes matricization of tensor \( \bm{ \mathcal{A} } \). Furthermore, given \( \bm{U} _{t} \), \( \bm{U} _{t_{l}} \), and \( \bm{U} _{t_{r}} \) as bases of the column spaces of \( \bm{A} ^{(t)} \), \( \bm{A} ^{(t_{l})} \), and \( \bm{A} ^{(t_{r})} \), respectively, we have
\begin{equation}\label{Eq_HT_Basis}
\bm{U} _{t} = (\bm{U} _{t_{l}} \otimes \bm{U} _{t_{r}})\bm{B} _{t}
\end{equation}
where \( \bm{U} _{t} \in \mathbb{R} ^{n_{t_{1}}n_{t_{2}} \cdots n_{t_{k}} \times r_{t}} \), and \( \bm{B} _{t} \in \mathbb{R} ^{r_{t_{l}}r_{t_{r}} \times r_{t}} \) is called transfer matrix. Note that \( r_{t} \), \( r_{t_{l}} \), and \( r_{t_{r}} \) are the respective ranks of \( \bm{A} ^{(t)} \), \( \bm{A} ^{(t_{l})} \), and \( \bm{A} ^{(t_{r})} \).

\begin{figure}
\centering
\includegraphics[width=0.4\textwidth]{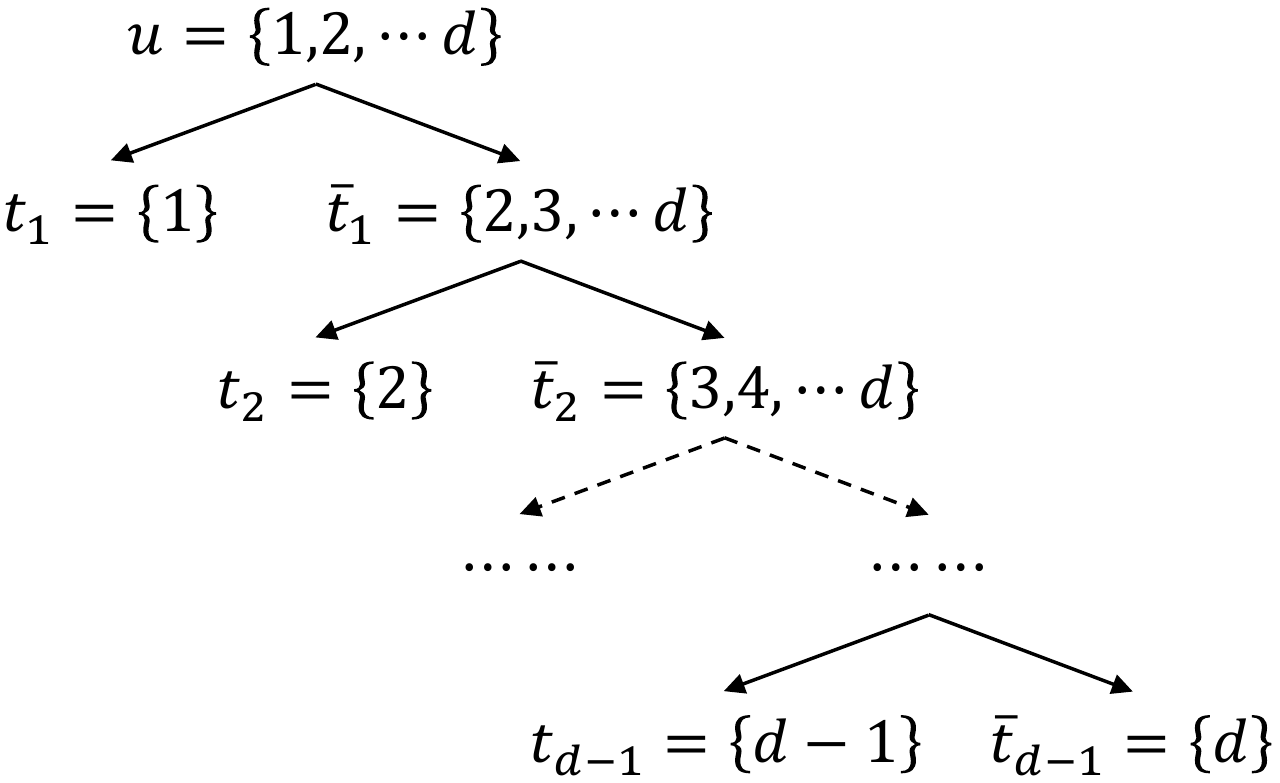}
\caption{\textbf{The degenerate dimension tree of tensor \(\bm{\mathcal{A}} \in \mathbb{R}^{n_1 \times n_2 \times \cdots \times n_d}\).}}
\label{Fig_degenerate_tree}
\end{figure}

It can be easily observed that only two subsets will be produced by utilizing Equation (\ref{Eq_HT_Basis}) each time. Thus, a specific hierarchical Tucker format is corresponding to a binary tree which continuously divides the original set \( u \) of full indices of modes until all the leaf nodes appear to be a singleton set of one index of mode. Such binary tree is called dimension tree, and there is a special form called degenerate dimension tree which splits out only one index of mode each time as shown in Figure \ref{Fig_degenerate_tree}.

According to Lemma 5.2 in \citet{Grasedyck_2010_InventHT} which proves that the hierarchical Tucker corresponding to the degenerate dimension tree yields the TT format. For the \(d\)th-order tensor in Figure \ref{Fig_degenerate_tree}, each entry of the \(q\)th (\(q \in \{2,3,\cdots,d-1\}\)) TT core tensor can be represented like
\begin{equation*}
\mathcal{G}_q(l,k,m) = \sum_{r=1}^{r_{\bar{t}_{q}}} \mathcal{B}_{\bar{t}_{q-1}}(l,r,m) U_{t_{q}}(k,r)
\end{equation*}
where \(t_{q} = \{q\}\), \(\bar{t}_{q} = \{q+1,\cdots,d\}\), \(\bm{U}_{t_{q}} \in \mathbb{R}^{n_{q} \times r_{\bar{t}_{q}}}\), \(\bm{\mathcal{B}}_{\bar{t}_{q-1}} \in \mathbb{R}^{r_{t_{q}} \times r_{\bar{t}_{q}} \times r_{t_{q-1}}}\), \(\bm{\mathcal{G}}_{q} \in \mathbb{R}^{r_{t_{q}} \times n_{q} \times r_{t_{q-1}}}\), and \(l,k,m,r\) are indices of the modes \(r_{t_{q}},n_{q},r_{t_{q-1}},r_{\bar{t}_{q}}\), respectively. Note that the implicit \(\bm{U}_{\bar{t}_{q}} \in \mathbb{R}^{n_{q+1}n_{q+2} \cdots n_{d} \times r_{t_{q}}}\) is defined based on Equation (\ref{Eq_HT_Basis}) in
\begin{equation*}
\bm{U}_{\bar{t}_{q-1}} = (\bm{U}_{t_{q}} \otimes \bm{U}_{\bar{t}_{q}}) \bm{B}_{\bar{t}_{q-1}}
\end{equation*}
where the transfer matrix \(\bm{B}_{\bar{t}_{q-1}} \in \mathbb{R}^{r_{t_{q}}r_{\bar{t}_{q}} \times r_{t_{q-1}}}\) is the initial form of tensor \(\bm{\mathcal{B}}_{\bar{t}_{q-1}}\).

Trace back to Equation (\ref{Eq_TTBaseTensor}), any TT rank of a \(d\)th-order tensor \(\bm{\mathcal{A}}\) is actually the \(r_{t_{q}}\) which comes from the matrix rank of \(\bm{A}^{(\bar{t}_{q})} \in \mathbb{R}^{n_{q+1}n_{q+2} \cdots n_{d} \times n_{1}n_{2} \cdots n_{q}}\) with the base of column space \(\bm{U}_{\bar{t}_{q}}\). For example, a 4th-order tensor \(\bm{\mathcal{X}} \in \mathbb{R}^{4 \times 5 \times 6 \times 7}\) whose serial \(t\)-modes matricizations based on the degenerate dimension tree are \(\bm{X}^{(5 \cdot 6 \cdot 7) \times 4}\), \(\bm{X}^{(6 \cdot 7) \times (4 \cdot 5)}\), and \(\bm{X}^{7 \times (4 \cdot 5 \cdot 6)}\), then the TT ranks should be \(r_1=4\), \(r_2=20\), and \(r_3=7\) if we suppose that these three matrices are all full rank. However, if higher value of rank is decided, the compression ratio \(R\) will be closer to the lower bound \(B(r^{(u)})\) which makes compression meaningless according to Inequality (\ref{Eq_CompressRatio_3DCNN}). Besides, higher values of TT ranks always cause ripples for numerical computation. In practice of DNNs, because of the considerable redundancy in weight matrices \citep{Denil_2013_Redundancy}, we can just consider the truncated TT format by selecting a comparatively small rank to replace all original ranks, e.g., set \(r_{2}=r_{1}=4\) and \(r_{3}=r_{1}=4\) for the above tensor \(\bm{\mathcal{X}}\), even the final compression ratio \(R\) should be very close to the upper bound \(B(r^{(l)})\). $\\$

\subsubsection{Pragmatic Verification}\quad

In order to verify the effectiveness of truncated TT format for training DNNs, we make two brief tests based on CIFAR-10 dataset \citep{Alex_2009_CIFAR10} to examine whether it is enough to just select a relatively small value of matrix rank from any \(\bm{A}^{(\bar{t}_{q})}\). Each test is executed through a CNN, and its TT CNN form compresses all but \(1 \times 1\) convolutional kernels based on Equation (\ref{Eq_TTCNN}). The detailed network architectures are shown in Table \ref{Table_choose_TT_ranks}. Furthermore, we force all the TT ranks to be the same value with range from 2 to 100 for easily observing the relationship between the accuracy and the value of rank.

\begin{table*}
\caption{\textbf{Design of CNNs for verifying the principle to select TT ranks.} Thereinto, each content in ``Conv'' and ``Shape'' denotes a convolutional kernel \(\bm{\mathcal{K}} \in \mathbb{R}^{l \times l \times C \times S}\) as \((l \times l) \times (C \times S)\), while its corresponding content in ``Conv'' and ``TT'' denotes the matching TT format from Equation (\ref{Eq_TTCNN}) as \((l \cdot l) \times (c_1 \cdot s_1) \times \cdots \times (c_d \cdot s_d)\).}
\label{Table_choose_TT_ranks}
  \centering
  \renewcommand\arraystretch{1.6}
  \resizebox{0.9\textwidth}{!}{
  \begin{tabular}{c | c c | c c}
   \hline
   \hline
     -- & \multicolumn{2}{c|}{CNN-1} & \multicolumn{2}{c}{CNN-2} \\
     \hline
     Layers & Shape & TT & Shape & TT \\
   \hline
   \hline
     Input & \(32 \times 32 \times 3\) & -- & \(32 \times 32 \times 3\) & -- \\
     Conv 1.1 & \((3 \times 3) \times (3 \times 64)\) & -- & \((3 \times 3) \times (3 \times 64)\) & -- \\
     Conv 1.2 & \((3 \times 3) \times (64 \times 64)\) & \((3 \cdot 3) \times (4 \cdot 4) \times (4 \cdot 4) \times (4 \cdot 4)\) & \((3 \times 3) \times (64 \times 64)\) & \((3 \cdot 3) \times (4 \cdot 4) \times (4 \cdot 4) \times (4 \cdot 4)\) \\
     Max Pooling 1 & \(2 \times 2\) & -- & \(2 \times 2\) & -- \\
     Conv 2.1 & \((3 \times 3) \times (64 \times 128)\) & \((3 \cdot 3) \times (4 \cdot 4) \times (4 \cdot 8) \times (4 \cdot 4)\) & \((1 \times 1) \times (64 \times 256)\) & -- \\
     Conv 2.2 & \((3 \times 3) \times (128 \times 128)\) & \((3 \cdot 3) \times (4 \cdot 8) \times (8 \cdot 4) \times (4 \cdot 4)\) & \((3 \times 3) \times (256 \times 256)\) & \((3 \cdot 3) \times (4 \cdot 4) \times (4 \cdot 4) \times (4 \cdot 4) \times (4 \cdot 4)\) \\
     Max Pooling 2 & \(2 \times 2\) & -- & \(2 \times 2\) & -- \\
     Conv 3.1 & \((3 \times 3) \times (128 \times 128)\) & \((3 \cdot 3) \times (4 \cdot 8) \times (8 \cdot 4) \times (4 \cdot 4)\) & \((3 \times 3) \times (256 \times 256)\) & \((3 \cdot 3) \times (4 \cdot 4) \times (4 \cdot 4) \times (4 \cdot 4) \times (4 \cdot 4)\) \\
     Conv 3.2 & \((3 \times 3) \times (128 \times 128)\) & \((3 \cdot 3) \times (4 \cdot 8) \times (8 \cdot 4) \times (4 \cdot 4)\) & \((3 \times 3) \times (256 \times 256)\) & \((3 \cdot 3) \times (4 \cdot 4) \times (4 \cdot 4) \times (4 \cdot 4) \times (4 \cdot 4)\) \\
     Average Pooling & \(4 \times 4\) & -- & \(4 \times 4\) & -- \\
     Linear & 128 & -- & 256 & -- \\
     Output & 10 & -- & 10 & -- \\
   \hline
   \hline
  \end{tabular}}
\end{table*}

\begin{figure}
\centering
\subfigure[CNN-1]{\includegraphics[width=0.35\textwidth]{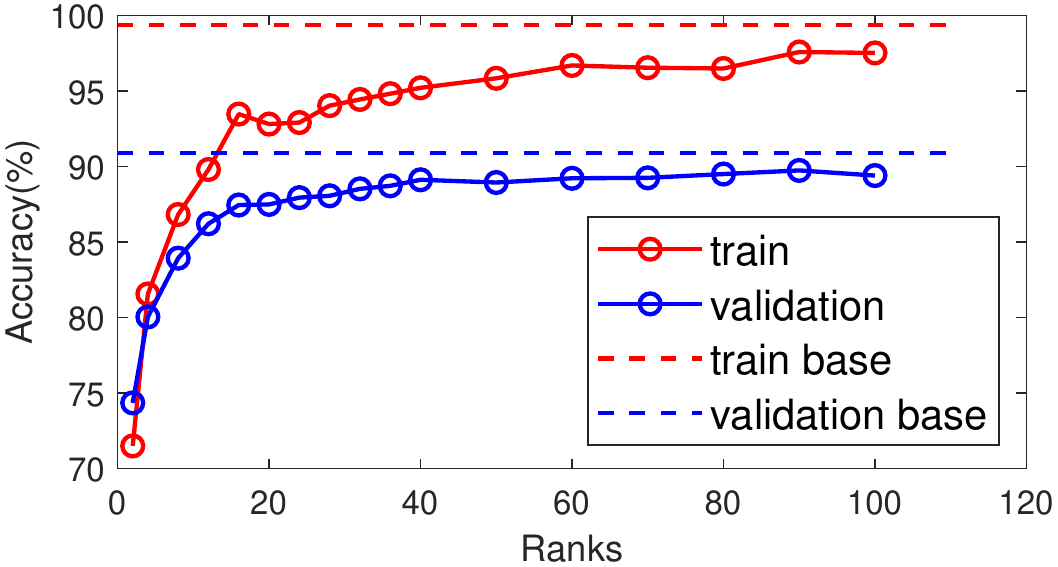}}
\subfigure[CNN-2]{\includegraphics[width=0.35\textwidth]{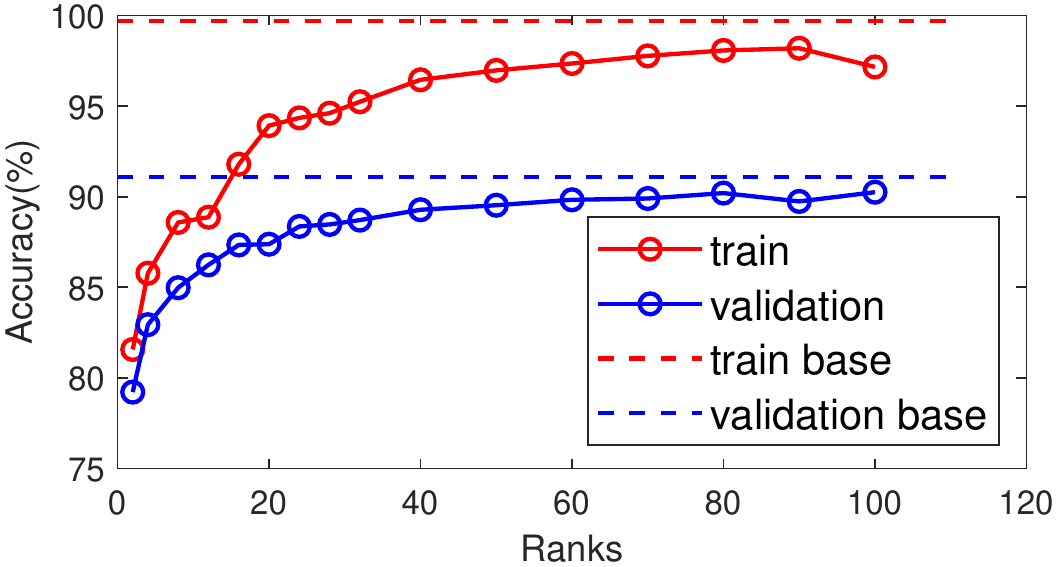}}
\caption{\textbf{Variation of accuracy with increasing the values of TT ranks.} Network architectures are defined in Table \ref{Table_choose_TT_ranks}. Thereinto, dash lines denote the performance of uncompressed networks.}
\label{Fig_tt_ranks_test}
\end{figure}

According to the definitions of TT format in Table \ref{Table_choose_TT_ranks}, the theoretical values of truncated TT ranks \(r_{t_{q}}\) should be 9, 16 or 32 in CNN-1, and 9 or 16 in CNN-2, respectively. Because \(r_{t_{q}}=16\) occupies the most proportion of all feasible values of original \(r_{t_{q}}\) in both CNN-1 and CNN-2, 16 could be the most suitable rank for truncated TT format with the premise that all values of TT ranks are the same. Moreover, the test results illustrated in Figure \ref{Fig_tt_ranks_test} show that the accuracy improvements have slowed down when passing the point at rank of 16 in both two TT CNNs.

In a nutshell, to design a truncated TT format for one layer in DNNs, we deem that for an arbitrary tensorizing \(d\)th-order weight or convolutional kernel tensor \(\bm{\mathcal{W}} \in \mathbb{R}^{n_1 \times n_2 \times \cdots \times n_d}\), one can select an appropriate rank \(r_1\) or \(r_{d-1}\) which is the full rank of \(\bm{W}^{(\bar{t}_{1})} \in \mathbb{R}^{n_{2}n_{3} \cdots n_{d} \times n_{1}}\) or \(\bm{W}^{(\bar{t}_{d-1})} \in \mathbb{R}^{n_{d} \times n_{1}n_{2} \cdots n_{d-1}}\), respectively. Approximately, in practice it is usually better to fine tune the \(i\)th TT rank \(r_i\) to be the full rank of \(\bm{W}^{(i)} \in \mathbb{R}^{n_i \times n_1 \cdots n_{i-1}n_{i+1} \cdots n_{d}}\). Although we have mentioned that making compression ratio \(R\) draw near to the upper bound \(B(r^{(l)})\) may cause accuracy loss according to Inequality (\ref{Eq_CompressRatio_3DCNN}), selecting truncated TT ranks is still sufficient for DNNs. 

\section{Experiments}\label{sec:Exp}

According to the inherent redundancy of DNNs \citep{Denil_2013_Redundancy}, we suppose that 3DCNNs are more redundant for easier compression, i.e., the accuracy loss of a TT compressed 3DCNN may be small or even taken away. Therefore, in this section, we first design five different 3DCNN models and their TT formats by continuously enlarging the network scale based on VIVA challenge dataset to observer how the accuracy loss is wiped out when the network redundancy increases. Second, under the limitation of our hardware, a carefully designed two stream 3DCNN and its TT format are trained on UCF11 and UCF101 datasets for further verification. All of our experiments are executed on TensorFlow and t3f library \citep{Novikov_2018_t3f}.

\subsection{Continuously Enlarged 3DCNNs}

\subsubsection{Dataset and Preprocessing}\quad

\begin{figure*}
\centering
\includegraphics[width=0.99\textwidth]{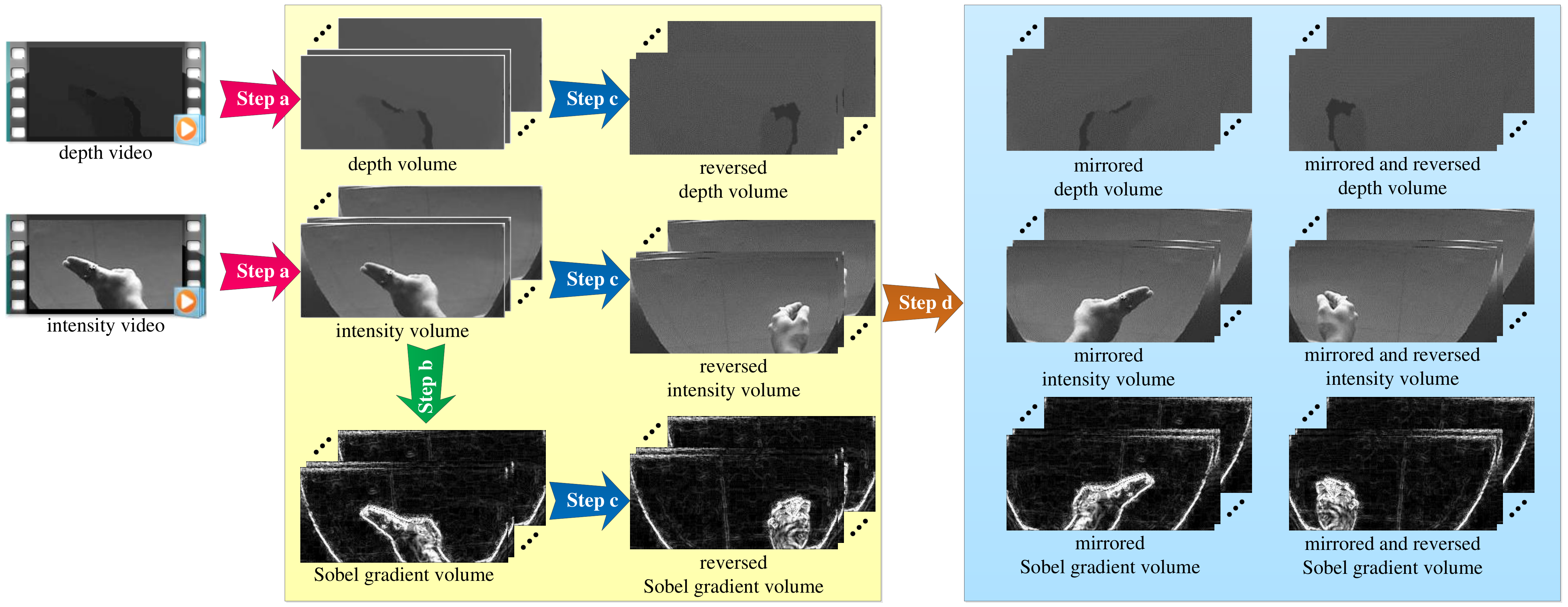}
\caption{\textbf{Data preprocessing for VIVA challenge dataset.} Note that not all samples can be processed in Step c because a minority of dynamic gestures do not have characteristic of time symmetry. All the frame images shown here are from the video with the serial number ``02\_01\_03'' in original dataset.}
\label{Fig_preprocess_VIVA}
\end{figure*}

We choose VIVA challenge dataset to observe whether the accuracy loss will reduce when scale of 3DCNN is growing, since it is a tiny but challenging hand gesture dataset \citep{Ohn-Bar_2014_VIVA,Molchanov_2015_3DCNN_1}, which is comprised of 885 video sequences including 19 dynamic hand gestures, and each video sequence contains two consistent channels that are intensity and depth.

Directly feeding the video into networks is doubtlessly naive, thus we mainly follow the existing data preprocessing strategy \citep{Molchanov_2015_3DCNN_1}, which increases the amount of samples, and the detailed steps are illustrated in Figure \ref{Fig_preprocess_VIVA}. Note that the nearest neighbor interpolation (NNI) \citep{Molchanov_2015_3DCNN_1} is used to normalize each video sequence to 32 frames, and we shrink every frame image to \(62 \times 28\) as width \(\times\) height by bicubic interpolation. Finally, every sample is a stacked frame sequence which has three channels, and we call each single channel as a volume, e.g., intensity volume, depth volume, or Sobel gradient volume.

\begin{figure}
\centering
\subfigure[Intensity]{
\begin{minipage}[b]{0.3\linewidth}
\includegraphics[width=1\linewidth]{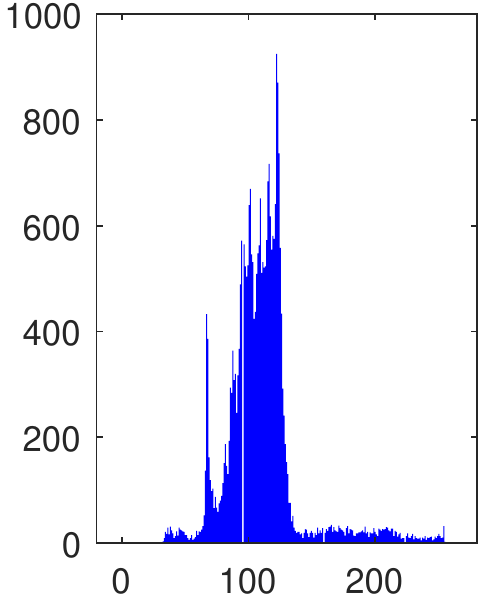}
\end{minipage}}
\subfigure[Depth]{
\begin{minipage}[b]{0.3\linewidth}
\includegraphics[width=1\linewidth]{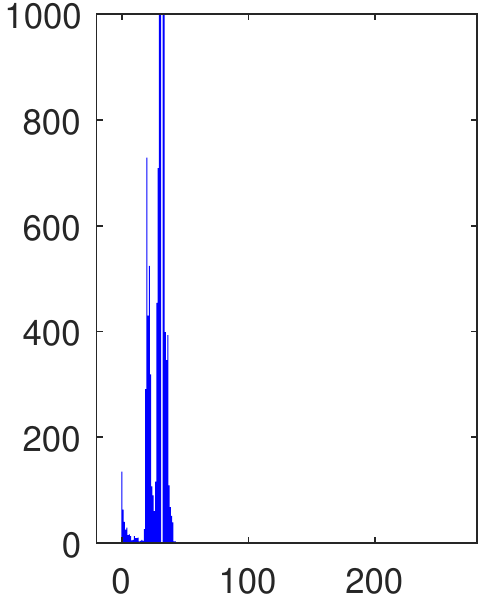}
\end{minipage}}
\subfigure[Sobel]{
\begin{minipage}[b]{0.3\linewidth}
\includegraphics[width=1\linewidth]{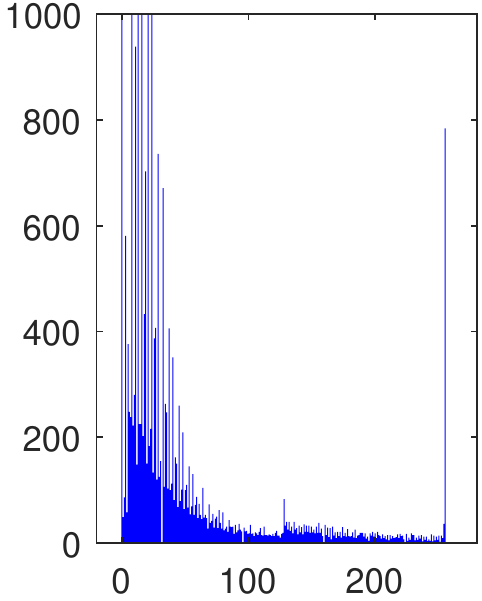}
\end{minipage}}
\caption{\textbf{Average gray histograms of frame images from volumes of different channels.} The corresponding video clip has the serial number ``01\_01\_01'' in original dataset.}
\label{Fig_hists}
\end{figure}

Besides, some data augmentation methods are also considered based on different characteristics of three channels, which can be observed by average gray histograms of different channels as shown in Figure \ref{Fig_hists}. Hence, except common affine transformation, we apply random contrast adjustment in the range of \(\pm 25\%\) for intensity frame images, increase the brightness randomly from \(0\% \sim 50\%\) for depth frame images, and drop out random \(50\%\) pixels (set their values to 0) for Sobel gradient frame images.

\subsubsection{Design of Networks}\quad

\begin{figure*}
\centering
\subfigure[Original networks]{\includegraphics[width=0.75\textwidth]{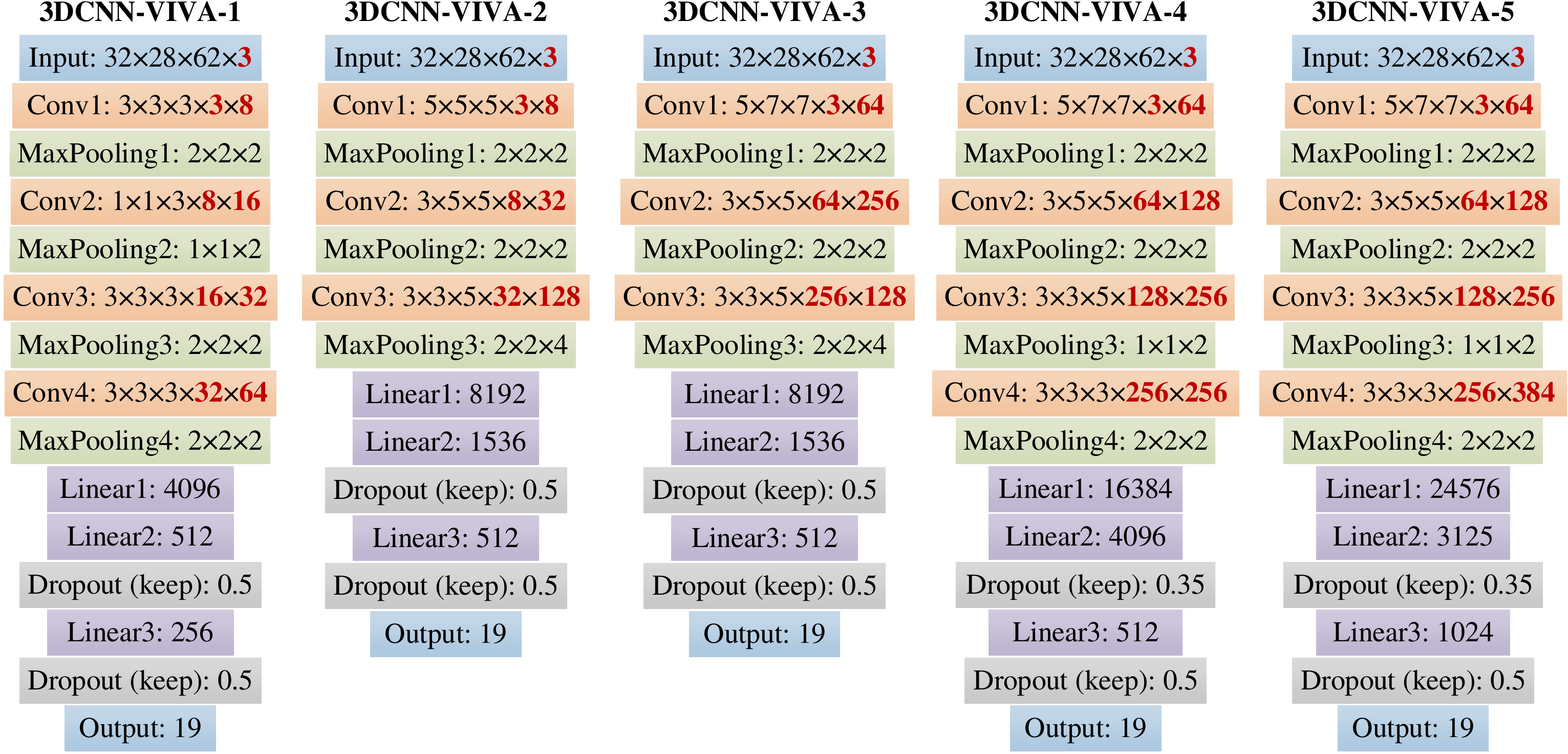}}
\subfigure[TT compressed networks]{\includegraphics[width=0.99\textwidth]{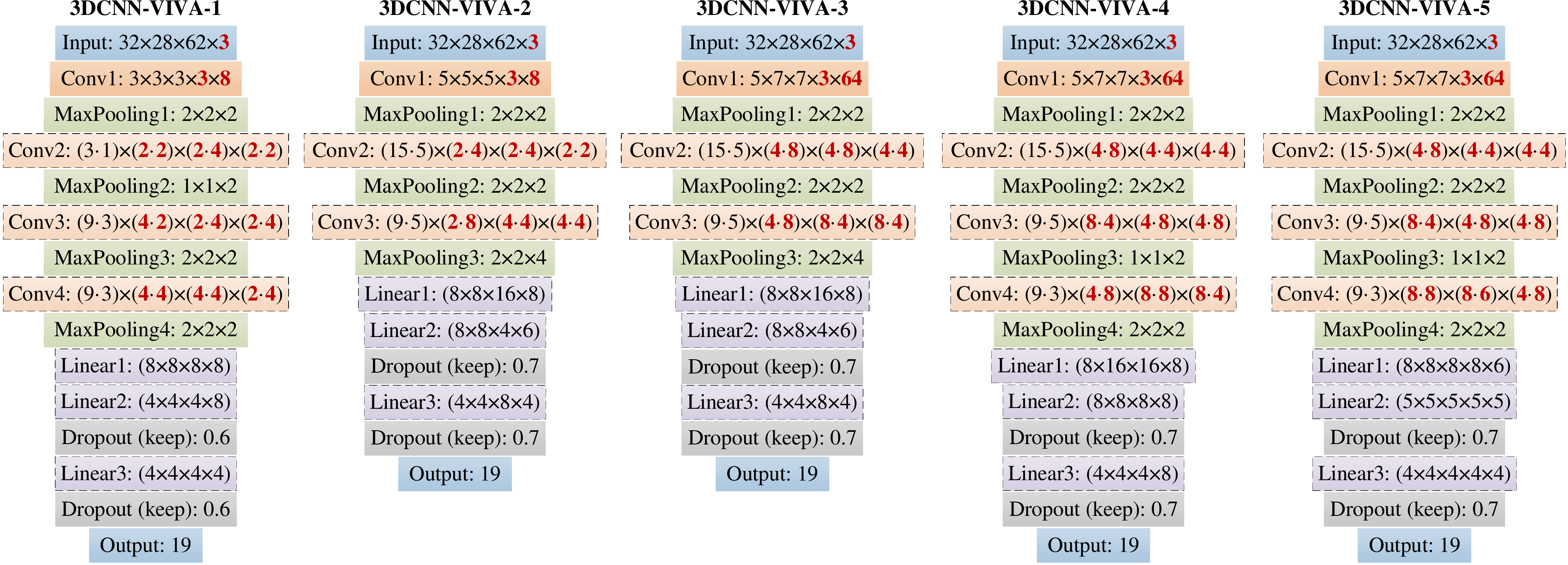}}
\caption{\textbf{Larger and larger network architectures for VIVA challenge dataset.} The numbers in red bold formats denote the modes of channels \(C\) or \(S\) in \(\mathcal{\bm{K}}_{3D} \in \mathbb{R}^{t \times h \times w \times C \times S}\) which represents the 3D convolutional kernel. Note that the number of neurons in every ``Linear1'' layer is equal to the number of output elements of corresponding last convolutional layer.}
\label{Fig_VIVA_3DCNNs}
\end{figure*}

In order to study the relationship between the compression performance and the network scale, as shown in Figure \ref{Fig_VIVA_3DCNNs}, we design five different 3DCNN models with larger and larger scale on VIVA challenge dataset. The detailed architectures are largely considered by referring to those in \citet{Molchanov_2015_3DCNN_1}, and we gradually enlarge the number of parameters in convolutional parts and whole networks from 3DCNN-VIVA-1 to 3DCNN-VIVA-5 as gathered in Table \ref{Table_VIVA_results}. We do not use batch normalization (BN) here to avoid its possible potential impact that may influence the continuous comparisons between original and compressed networks throughout these five 3DCNNs. We select the comparatively small \(t\)-modes matricization ranks for the truncated TT decomposition. For instance, the tensorizing format of Conv2 layer in 3DCNN-VIVA-3 from Figure \ref{Fig_VIVA_3DCNNs} is \((15 \cdot 5) \times (4 \cdot 8) \times (4 \cdot 8) \times (4 \cdot 4)\) which yields a 4th-order tensor with the shape of \(75 \times 32 \times 32 \times 16\), then we let 16 to be the value of all TT ranks. Furthermore, it is important to make the keeping probability (dropout parameter) of TT compressed network higher than the uncompressed ones because the TT FC layers have much less neurons than the original FC layers. Finally, all the activation functions omitted in Figure \ref{Fig_VIVA_3DCNNs} are ReLU.

\subsubsection{Learning and Results}\quad

We consider the k-fold cross validation that split 3100 samples (after preprocessing) into 2500 samples as training set and the rest 600 samples as validation set, and train totally 100 epochs for all networks. On account of the varying scale of different networks and multiple attempts, the optimal hyper-parameters for all networks are not very same, but we keep the paired original and TT compressed networks under the same condition. The initial learning rate is set to 0.01 for 3DCNN-VIVA-1, 3DCNN-VIVA-2 and 3DCNN-VIVA-3, 0.005 for 3DCNN-VIVA-4 and 3DCNN-VIVA-5, respectively. The learning rate decreases by a factor of 0.1 after every 30 epochs and the momentum is set to 0.9. Due to the limitation of our GPU resources, the mini-batch size is carefully designed for each network, and concretely, the batch size is 100 for 3DCNN-VIVA-1 and 3DCNN-VIVA-2, 50 for 3DCNN-VIVA-3, 25 for 3DCNN-VIVA-4 and 3DCNN-VIVA-5. Besides, we use stochastic gradient descent (SGD) to optimize our networks and the loss function is cross entropy with the softmax layer.

The results of these experiments about networks above are shown in Table \ref{Table_VIVA_results}, in which their storage requirements (``*.data'' file in TensorFlow) and the amounts of parameters (proportions of convolutional parts are highlighted) are also listed. One can easily observe that the accuracy loss, which is also termed as \emph{degeneration} in the table, decreases as the scale of each network increases in turn until it is vanished. Compression ratio, which is calculated based on the amount of whole parameters, is promoted from 11.5\(\times\) to 180.4\(\times\) that implies better compression performance can be easier obtained for redundant networks. Further detailed discussions are given in the next section.

\begin{table*}
\caption{\textbf{Experimental results on VIVA challenge dataset.} Note that the unit of degeneration ``pp'' means the percentage point and every decimal of degeneration is the difference value between the accuracy of the original network and its TT counterpart. ``Conv'' and ``Whole'' mean the number of parameters of convolutional part and whole network respectively.}
\label{Table_VIVA_results}
  \centering
  \renewcommand\arraystretch{1.6}
  \resizebox{0.98\textwidth}{!}{
  \begin{tabular}{c | c c | c c | c c | c c | c c}
   \hline
   \hline
     -- & \multicolumn{2}{c|}{3DCNN-VIVA-1} & \multicolumn{2}{c|}{3DCNN-VIVA-2} & \multicolumn{2}{c|}{3DCNN-VIVA-3} & \multicolumn{2}{c|}{3DCNN-VIVA-4} & \multicolumn{2}{c}{3DCNN-VIVA-5} \\
     \hline
     -- & Original & TT & Original & TT & Original & TT & Original & TT & Original & TT\\
   \hline
   \hline
     Accuracy (\%) & \begin{tabular}{c} 78.61 \\ \(\pm\) 2.05 \end{tabular} & \begin{tabular}{c} 71.75 \\ \(\pm\) 1.46 \end{tabular} & \begin{tabular}{c} 80.12 \\ \(\pm\) 2.01 \end{tabular} & \begin{tabular}{c} 74.28 \\ \(\pm\) 2.17 \end{tabular} & \begin{tabular}{c} 81.13 \\ \(\pm\) 2.01 \end{tabular} & \begin{tabular}{c} 77.67 \\ \(\pm\) 1.75 \end{tabular} & \begin{tabular}{c} 80.20 \\ \(\pm\) 2.85 \end{tabular} & \begin{tabular}{c} 80.78 \\ \(\pm\) 2.61 \end{tabular} & \begin{tabular}{c} 81.47 \\ \(\pm\) 1.19 \end{tabular} & \begin{tabular}{c} 81.83 \\ \(\pm\) 0.76 \end{tabular}\\
     Degeneration (pp) & \multicolumn{2}{c|}{6.86} & \multicolumn{2}{c|}{5.84} & \multicolumn{2}{c|}{3.46} & \multicolumn{2}{c|}{-0.58} & \multicolumn{2}{c}{-0.36} \\
     Storage (MB) & 26.3 & 2.23 & 155 & 3.29 & 184 & 5.28 & 836 & 15.6 & 970 & 5.41 \\
     \begin{tabular}{c} Parameters \\ Conv/Whole \end{tabular} (\(10^6\)) & 0.07/2.3 & 0.015/0.2 & 0.2/13.6 & 0.015/0.29 & 2.75/16.03 & 0.13/0.46 & 3.9/73.12 & 0.23/1.36 & 4.79/84.8 & 0.25/0.47 \\
     Compression Ratio & \multicolumn{2}{c|}{11.5\(\times\)} & \multicolumn{2}{c|}{46.9\(\times\)} & \multicolumn{2}{c|}{34.85\(\times\)} & \multicolumn{2}{c|}{53.8\(\times\)} & \multicolumn{2}{c}{180.4\(\times\)} \\
   \hline
   \hline
  \end{tabular}}
\end{table*}

\subsection{Two Stream 3DCNN for Verification}

\subsubsection{Dataset and Preprocessing}\quad

\begin{figure}
\centering
\subfigure[VIVA]{
\begin{minipage}[b]{0.4\linewidth}
\includegraphics[width=1\linewidth]{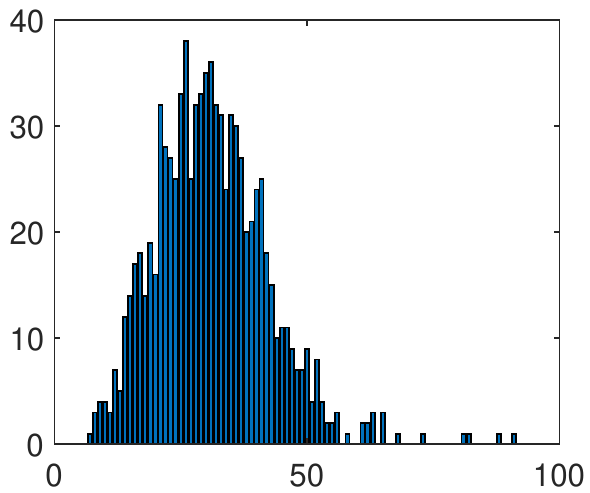}
\end{minipage}}
\subfigure[UCF11]{
\begin{minipage}[b]{0.4\linewidth}
\includegraphics[width=1\linewidth]{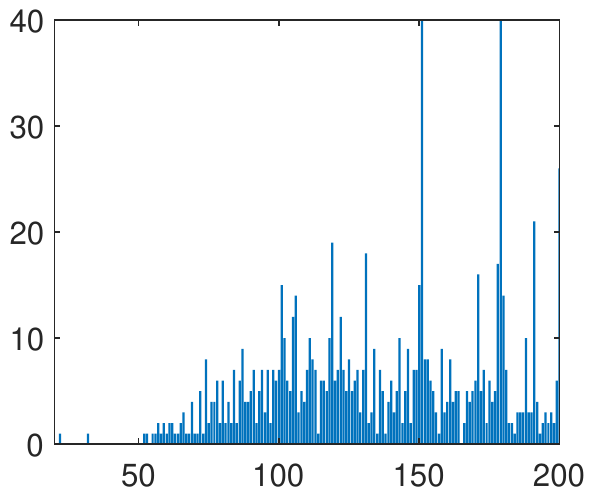}
\end{minipage}}
\caption{\textbf{Statistical histograms of number of frames of VIVA and UCF11 datasets.} The horizontal axis denotes the number of frames and the vertical axis denotes the amount of videos.}
\label{Fig_statistics}
\end{figure}

The foregoing experiments on VIVA challenge dataset imply that the redundant enough 3DCNNs may be easier to be compressed even losslessly. Here we choose two widely used benchmark datasets, i.e., UCF11 and UCF101, and design a redundant enough (limited by our hardware) two stream 3DCNN to experiment based on RGB and optical flow frames. The final frame size to be fed is 80 \(\times\) 60 as weight \(\times\) height, and the optical flow data is calculated by Farneback algorithm \citep{Farneback_2003_OpticalFlow}.

Whether UCF11 or UCF101 has more complex and flat distribution of video length than VIVA as shown in Figure \ref{Fig_statistics}, thus NNI cannot be considered right along. Here we choose the random clipping \citep{Varol_2018_LongTerm3DCNN,Simonyan_2014_TwoStream}, which samples a consecutive frame sequence with a fixed length (50 in our experiments), for sampling training datasets. Particularly, since UCF101 is very easier to fall into over-fitting during training \citep{Hara_2018_Res3DCNN}, following \citet{Varol_2018_LongTerm3DCNN} we down sample the original frame from 320 \(\times\) 240 to 123 \(\times\) 92, and further randomly extract the frame with the size of 80 \(\times\) 60. Other data augmentation approaches we considered includes random affine transformation for both RGB and optical flow data, contrast and saturation adjustment for RGB frames only.

\subsubsection{Design of Networks}\quad

\begin{figure}
\centering
\subfigure[Original network]{\includegraphics[width=0.3\textwidth]{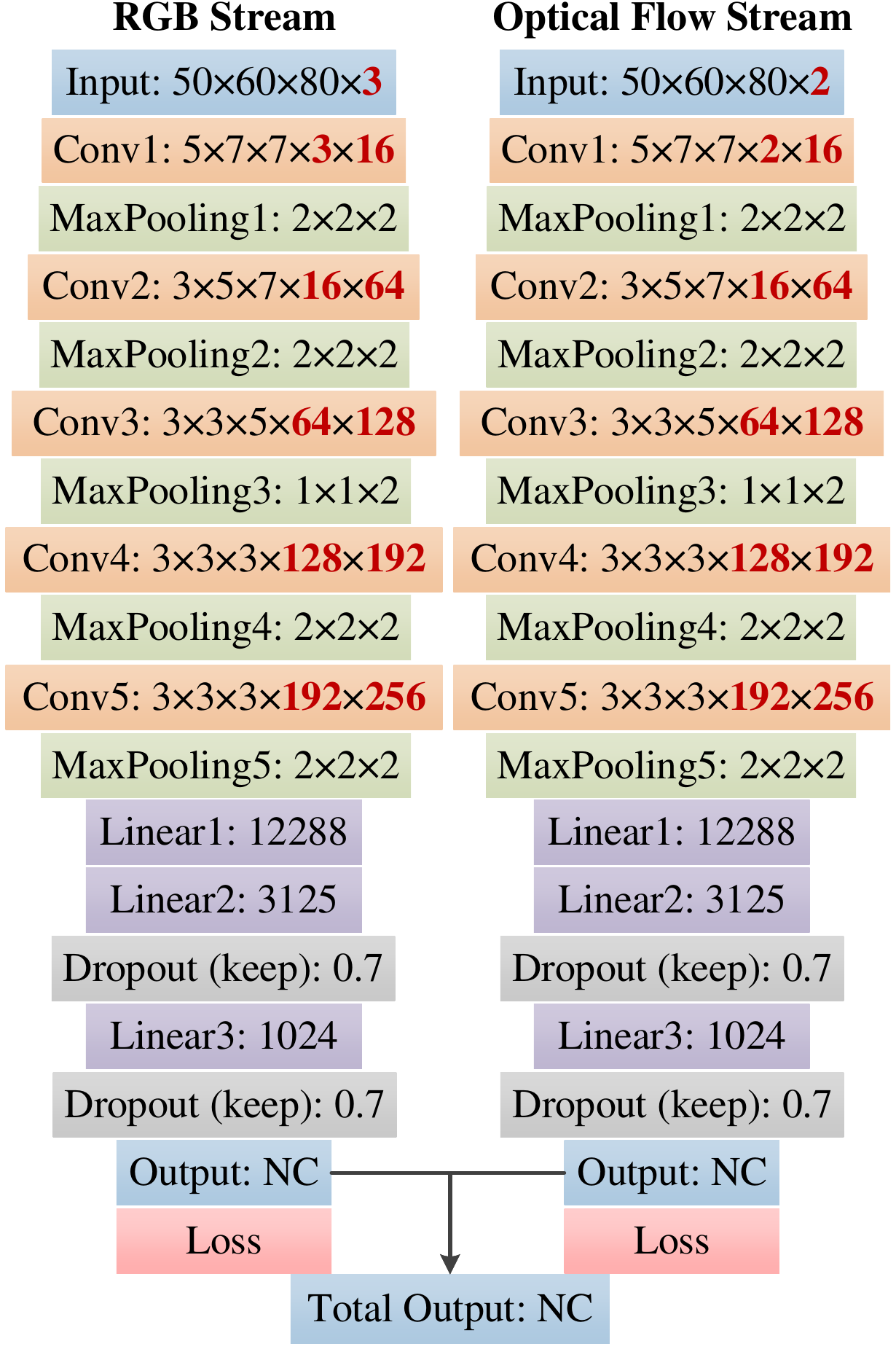}}
\subfigure[TT compressed network]{\includegraphics[width=0.4\textwidth]{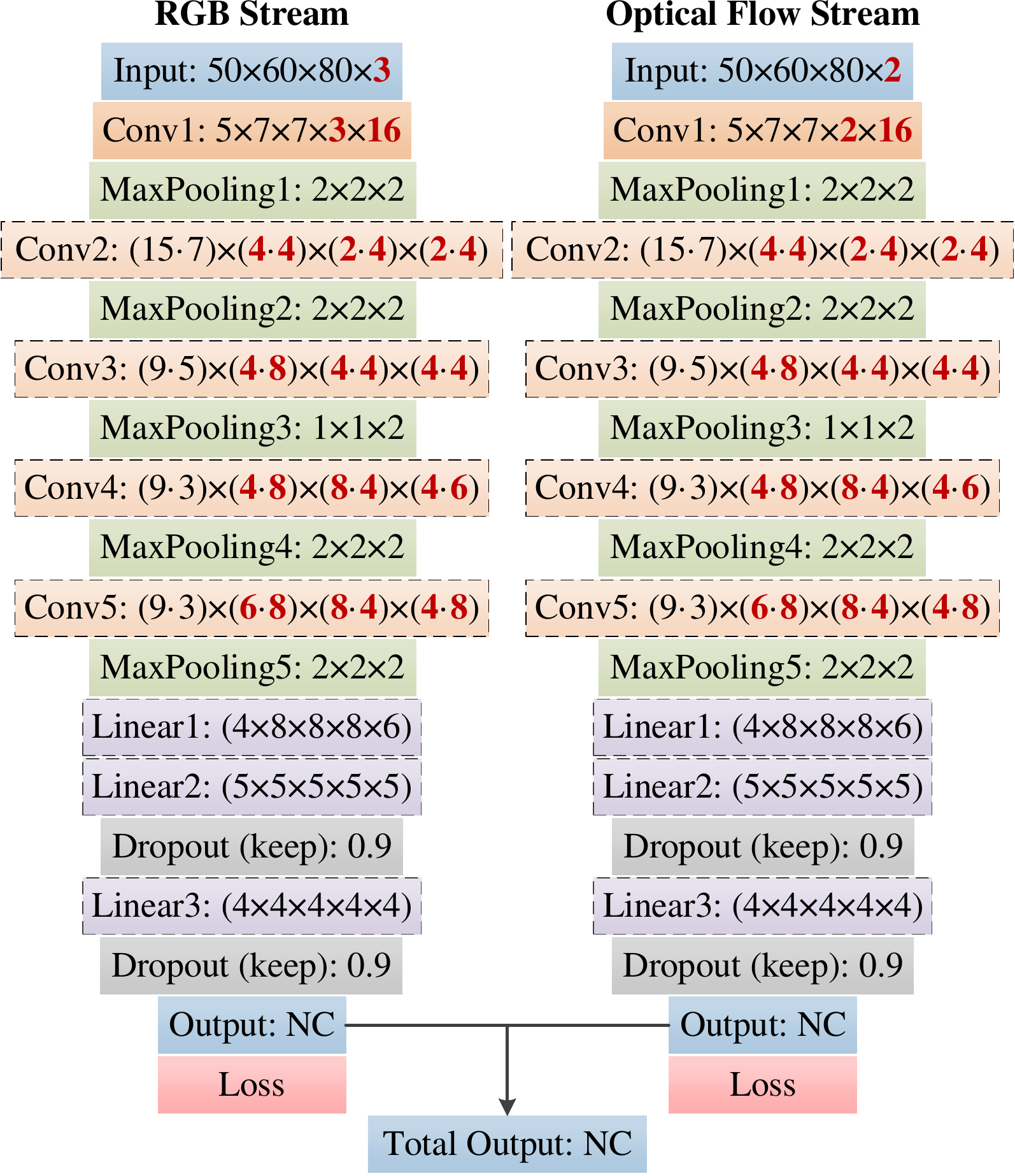}}
\caption{\textbf{Network architecture for UCF11 and UCF101 datasets.} The abbrev ``NC'' equals 11 or 101 for UCF11 or UCF101 respectively.}
\label{Fig_UCF_3DCNNs}
\end{figure}

The architecture of two stream 3DCNN is shown in Figure \ref{Fig_UCF_3DCNNs}, which mostly follows the 8 layers network in \citet{Varol_2018_LongTerm3DCNN} but has some adjustment of channels and filter size. Here BN is used ahead of activation function which is ReLU, since UCF datasets are more complex and we intend to check whether BN will influence the comparison between original and TT compressed networks.

\subsubsection{Learning and Results}\quad

There are three official splits in UCF101 dataset, thus we construct training and validation sets following every split and learn several times on them. For UCF11, both of the existing k-fold \citep{Yang_2017_TTRNN} and leave-one-group (LOG) \citep{Peng_2014_UCF11} cross validation are considered in our experiments. The initial learning rate is 0.003, the total number of training epochs is up to 100, and the learning rate decreases exponentially by 0.1 factor after every 30 epochs. We also use SGD optimizer and the momentum coefficient is set to 0.9. The batch size is 20 for both original and TT networks.

\begin{figure}
\centering
\includegraphics[width=0.48\textwidth]{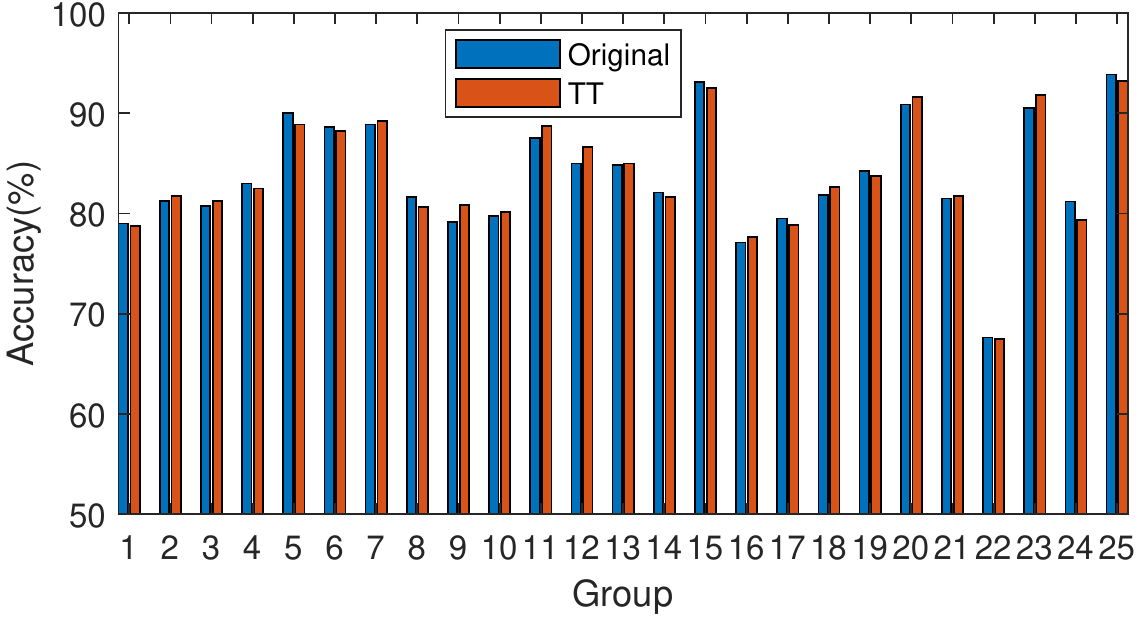}
\caption{\textbf{Experimental results on UCF11 dataset (LOG).} The horizontal axis represents which group is decided to be the validation set.}
\label{Fig_UCF11_LOG}
\end{figure}

The results are shown in Table \ref{Table_UCF_results}, among which the storage consumption of original 3DCNN is huge (nearly 1 GB), thus the compression ratio is considerable (around one hundred times). Fortunately, these results on UCF datasets verify that our TT compressed 3DCNN can still keep performance (even a little degeneration occurs on UCF101), which is hard to be implemented in 2DCNNs \citep{Garipov_2016_TTCNN}. By observing the variation of parameters, both convolutional and FC parts are compressed heavily, and the proportion of convolutional parameters is promoted. The tiny difference of space complexity between UCF11 and UCF101 is caused by the last FC layer whose dimension of output is decided by the number of classes. Besides, it should be emphasized that we only record the average accuracy on UCF11 (LOG) because the distribution of validation accuracy for different groups has significant variation, which can be obviously noticed in Figure \ref{Fig_UCF11_LOG}.

\begin{table}
\caption{\textbf{Experimental results on UCF11 and UCF101 datasets.}}
\label{Table_UCF_results}
  \centering
  \renewcommand\arraystretch{1.6}
  \resizebox{0.45\textwidth}{!}{
  \begin{tabular}{c | c c}
   \hline
   \hline
     -- & Original & TT \\
   \hline
   \hline
     Accuracy (\%) \begin{tabular}{c} UCF11 (k-fold) \\ UCF11 (LOG) \\ UCF101 \end{tabular} & \begin{tabular}{c} 93.78 \(\pm\) 1.09 \\ 83.72 \\ 78.38 \(\pm\) 0.37 \end{tabular} & \begin{tabular}{c} 93.56 \(\pm\) 1.16 \\ 83.81 \\ 77.86 \(\pm\) 0.51 \end{tabular} \\
   \hline
     Storage (MB) \begin{tabular}{c} UCF11 \\ UCF101 \end{tabular} & \begin{tabular}{c} 1009.57 \\ 1011.68 \end{tabular} & \begin{tabular}{c} 9.9 \\ 12 \end{tabular} \\
   \hline
     \begin{tabular}{c} Parameters \\ Conv/Whole \end{tabular} (\(10^6\)) \begin{tabular}{c} UCF11 \\ UCF101 \end{tabular} & \begin{tabular}{c} 4.95/88.17 \\ 4.95/88.36 \end{tabular} & \begin{tabular}{c} 0.36/0.82 \\ 0.36/1.01 \end{tabular} \\
   \hline
     Compression Ratio \begin{tabular}{c} UCF11 \\ UCF101 \end{tabular} & \multicolumn{2}{c}{\begin{tabular}{c} 107.5\(\times\) \\ 87.5\(\times\) \end{tabular}} \\
   \hline
   \hline
  \end{tabular}}
\end{table}

\subsection{Comparison with Typical Works}

For making our experiments convinced and basic for the following discussions, here we make a comparison between our results and recent typical practices on VIVA challenge, UCF11 and UCF101 datasets in Table \ref{Table_Comp_Typical_Works}. It can be confirmed that the results of our experiments can be comparable to corresponding state-of-the-art practices. More importantly, since there are not any significant accuracy losses for our TT networks, we can say that the TT compressed 3DCNNs have ability to compete with the state-of-the-arts at least on VIVA challenge, UCF11 and UCF101 datasets.

Notice that since training 3DCNNs from scratch on UCF\-101 is difficult \citep{Hara_2018_Res3DCNN} and we concern more about \emph{in situ} training of TT compressed 3DCNNs in this work, we only list intermediate results which are trained from scratch by corresponding typical works. Similarly, results produced by some pre-trained approaches on UCF11 are not listed either, e.g., pre-trained CNNs + LSTM in \citet{Pan_2019_UCF11}.

\begin{table}
\caption{\textbf{Comparing our experimental results with typical works on VIVA challenge, UCF11 and UCF101 datasets.}}
\label{Table_Comp_Typical_Works}
  \centering
  \renewcommand\arraystretch{1.6}
  \resizebox{0.45\textwidth}{!}{
  \begin{tabular}{c | c}
   \hline
   \hline
     \multicolumn{2}{c}{VIVA challenge} \\
   \hline
     Method & Accuracy (\%) \\
   \hline
   \hline
     Dense Trajectories \citep{Wang_2013_DenseTrajectories} & 54 \\
     HON4D \citep{Oreifej_2013_HON4D} & 58.7 \\
     HOG + HOG\(^{2}\) \citep{Ohn-Bar_2014_VIVA} & 64.5 \\
     Multi-resolution 3DCNNs \citep{Molchanov_2015_3DCNN_1} & 77.5 \\
     Ours Original/TT & \textbf{81.47/81.83} \\
   \hline
   \hline
     \multicolumn{2}{c}{UCF11} \\
   \hline
     Method & Accuracy (\%) \\
   \hline
   \hline
     Fisher Vector (LOG) \citep{Peng_2014_UCF11} & 93.77 \\
     Local Motion (LOG) \cite{Cho_2014_UCF11} & 88.0 \\
     Visual Attention (k-fold) \citep{Sharma_2016_UCF11} & 85.0 \\
     TT-GRU (k-fold) \citep{Yang_2017_TTRNN} & 81.3 \\
     BT-LSTM (k-fold) \citep{Ye_2018_UCF11} & 85.3 \\
     TR-LSTM (k-fold) \citep{Pan_2019_UCF11} & 86.9 \\
     Ours Original/TT (LOG) & \textbf{83.72/83.81} \\
     Ours Original/TT (k-fold) & \textbf{93.78/93.56} \\
   \hline
   \hline
     \multicolumn{2}{c}{UCF101 (Training from scratch)} \\
   \hline
     Method & Accuracy (\%) \\
   \hline
   \hline
     C3D \citep{Tran_2015_3DCNN} & 44 \\
     3DConvNet \citep{Carreira_2017_UCF101} & 79.9 \\
     3D ResNet-18 \citep{Hara_2018_Res3DCNN} & 42.4 \\
     MiCT \citep{Zhou_2018_UCF101} & 58.7 \\
     LTC \citep{Varol_2018_LongTerm3DCNN} & 80.5 \\
     Ours Original/TT & \textbf{78.38/77.86} \\
   \hline
   \hline
  \end{tabular}}
\end{table}

\section{Discussions}\label{sec:Dis}

Any potential of compression is derived from the inherent redundancy of DNNs \citep{Denil_2013_Redundancy}, and we are aware of that 3DCNNs have higher level of redundancy than traditional normal 2DCNNs in the light of our experiments and such phenomenon can permit us to develop a low less compression method based on TT decomposition. Therefore, in this section, we will emphasize more on the redundancy of 3DCNNs. Besides, some other aspects, e.g., regularization, latent degradation, computation complexity, and core significance of TT, shall also be analyzed.

\subsection{Redundancy of 3DCNNs}

\subsubsection{Large Convolutional Kernel Size}\quad

A stack of two 3\(\times\)3 convolutional kernels with fewer parameters has the equivalent receptive field as a single 5\(\times\)5 convolutional kernel which is realized in VGG-Net \citep{Simonyan_2015_VGG}, and the principle of using 3\(\times\)3 kernel is widely adopted nowadays \citep{He_2016_ResNet,Huang_2017_DenseNet}. Furthermore, \citet{Tran_2015_3DCNN} also claim that 3\(\times\)3\(\times\)3 kernel in 3DCNNs is the best. However, 3\(\times\)3\(\times\)3 convolutional kernel is unfriendly for TT decomposition, since the stacked tiny convolutional kernels is inherently a design of compact architecture which can reduce the redundancy. That may be the reason why compressing classical CNNs with 3\(\times\)3 kernels is hard to avoid accuracy loss \citep{Garipov_2016_TTCNN}. Some examples are described below for further explanation.

From network 3DCNN-VIVA-1 to network 3DCNN-VIVA-3, the kernel sizes and channels increase gradually without deepening the network, and the accuracy degeneration decreases accordingly. In contrast, we try to redesign the kernels in 3DCNN-VIVA-2 as 3\(\times\)3, i.e., transform the Conv1 from \(5 \times 5 \times 5\) to a stack of two \(3 \times 3 \times 3\) layers, transform the Conv2 from \(3 \times 5 \times 5\) to a stack of \(3 \times 3 \times 3\) and \(1 \times 3 \times 3\) layers, and transform the Conv3 from \(3 \times 3 \times 5\) to a stack of \(3 \times 3 \times 3\) and \(1 \times 1 \times 3\) layers. As we do so, the degeneration increases even the performance of both the uncompressed and TT networks increase concurrently. Similarly, for our two stream 3DCNN in Figure \ref{Fig_UCF_3DCNNs} on UCF11 (k-fold), if all the kernel sizes are replaced by \(3 \times 3 \times 3\), degeneration will occur (the accuracy of TT network is around 92.19\%) and the accuracy of original network has no evident variation.

\subsubsection{Appropriate Number of Channels}\quad

In CNNs, more channels represent more possible feature combinations, which can positively improve the performance of networks. The convolution in TT format in  Equation (\ref{Eq_TTCNN}) or Equation (\ref{Eq_TT_3DCNN}) allows us to design wider CNNs under the restriction of storage capacity. The performance of 3DCNN-VIVA-3 verifies this by comparing with 3DCNN-VIVA-2, in which the former network has higher proportion of convolutional parameters shown in Table \ref{Table_VIVA_results}.

Furthermore, more channels contribute to more balanced TT shapes that may bring higher compression ratio and better keeping information, particularly for FC layers \citep{Novikov_2015_TT}. Comparing 3DCNN-VIV-A-4 with 3DCNN-VIVA-5 in Table \ref{Table_VIVA_results}, the latter with higher compression ratio still has comparable performance with the former, by just enlarging the final output channels from 256 to 384. It is obvious in Figure \ref{Fig_VIVA_3DCNNs} that the shapes of (\(8 \times 8 \times 8 \times 8 \times 6\)) and (\(4 \times 4 \times 4 \times 4 \times 4\)) in 3DCNN-VIVA-5 are better than the corresponding shapes of (\(8 \times 16 \times 16 \times 8\)) and (\(4 \times 4 \times 4 \times 8\)) in 3DCNN-VIVA-4.

\subsubsection{Layer Coupling}\quad

It is necessary to point out that there exists coupling between convolutional and FC layers that affects the performance of TT CNNs. \citet{Novikov_2015_TT,Garipov_2016_TTCNN} show that compressing the FC part may get sufficient compression ratio with a little accuracy loss, while compressing both FC and convolutional parts is still hard to avoid degeneration. Thus, it seems that compressing the convolutional part is unnecessary. However, we find the coupling between convolutional layers and FC layers gives the meaning to compress convolutional kernels.

In detail, for the network 3DCNN-VIVA-1, we severally compress the FC part, the convolutional part, and both of these two parts. The results in Table \ref{Table_coupling} indicate that just compressing either part can cause considerable accuracy loss, while compressing both parts has not produced more substantial degeneration. Furthermore, compressing the whole network can improve the compression ratio obviously (from 9.2\(\times\) to 11.5\(\times\)) comparing with compressing FC part only. In a word, compressing the whole networks especially for 3DCNNs will not be worse than compressing either layer part (convolutional or FC).

On the other hand, when the scale of 3DCNN grows, amount of parameters in convolutional part appears to be more considerable, e.g., 3DCNN-VIVA-5 and two stream 3DCNN have \(4.79 \times 10^6\) and \(4.95 \times 10^6\) convolutional parameters according to Table \ref{Table_VIVA_results} and \ref{Table_UCF_results} respectively, so compressing convolutional kernels is necessary. Therefore, by taking into account both the layer coupling and the considerable amount of convolutional parameters in large scaled 3DCNNs, we deem that discussing the redundancy of 3DCNNs should focus on the entire network.

\begin{table}
\caption{\textbf{Compressing different parts of 3DCNN-VIVA-1.} The ``Base'' column denotes the uncompressed network.}
\label{Table_coupling}
  \centering
  \renewcommand\arraystretch{1.6}
  \resizebox{0.48\textwidth}{!}{
  \begin{tabular}{c | c c c c}
    \hline
    \hline
    Compressing Part & Base & FC & Convolution & Both \\
    \hline
    \hline
    Accuracy (\%) & \begin{tabular}{c} 78.61 \\ \(\pm 2.05\) \end{tabular} & \begin{tabular}{c} 72.97 \\ \(\pm 1.78\) \end{tabular} & \begin{tabular}{c} 72.58 \\ \(\pm 0.68\) \end{tabular} & \begin{tabular}{c} 71.75 \\ \(\pm 1.46\) \end{tabular} \\
    \hline
    Degeneration (pp) & - & 5.64 & 6.03 & 6.86 \\
    \hline
    \begin{tabular}{c} Parameters \\ Conv/Whole \end{tabular} (\(10^6\)) & 0.07/2.3 & 0.07/0.25 & 0.015/2.25 & 0.015/0.2 \\
    \hline
    Compression Ratio & - & 9.2\(\times\) & 1.02\(\times\) & 11.5\(\times\) \\
    \hline
    \hline
  \end{tabular}}
\end{table}

\subsubsection{Scale of Entire Networks}\quad

Regarding the network scale, we find that the larger network may be easier to be compressed. Comparing the performance of network 3DCNN-VIVA-4 and network 3DCNN-VIVA-5 in Table \ref{Table_VIVA_results}, the latter can obtain better accuracy with even lower storage cost. We believe that the larger 3DCNNs contain stronger redundancy.

\begin{figure}
\centering
\includegraphics[width=0.4\textwidth]{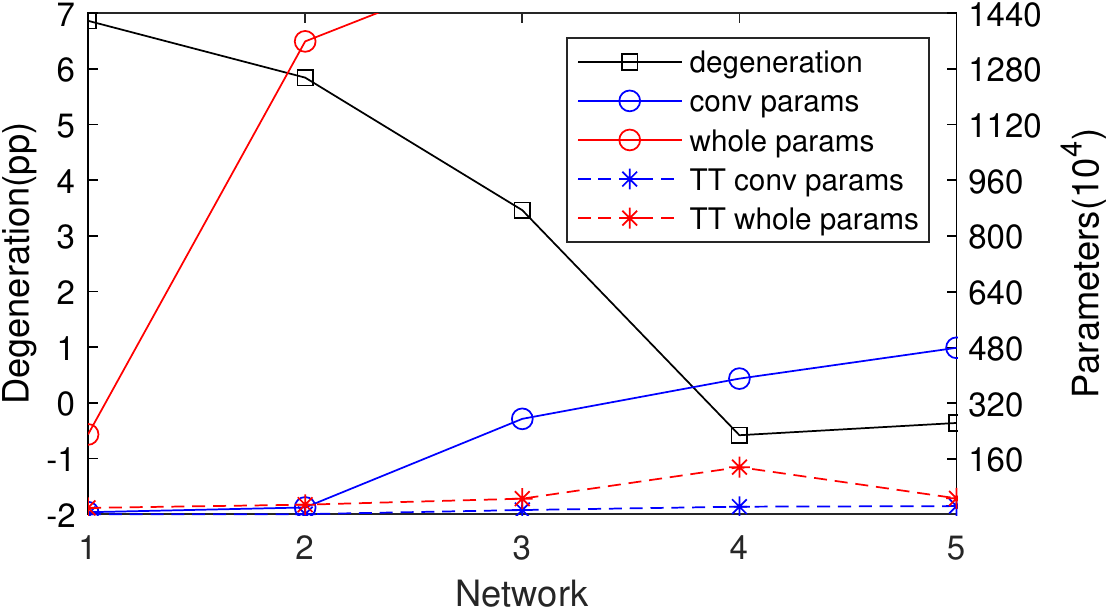}
\caption{\textbf{Variations of parameter amount and accuracy degeneration of the networks on VIVA challenge dataset.} Note that the number of whole parameters of uncompressed networks is too large to draw in this figure.}
\label{Fig_scale}
\end{figure}

Moreover, in general, a large and sparse DNN can get better performance than the dense one with same network scale, which phenomenon is also concluded by \citet{Zhu_2018_PruneOrNot}. This rule can be perceived in Figure \ref{Fig_scale} which illustrates the variations of parameters and accuracy degeneration from 3DCNN-VIVA-1 to 3DCNN-VIVA-5 according to Table \ref{Table_VIVA_results}. It can be sensed that, not only the amount of whole parameters, but also that of convolutional parameters, can be extremely reduced when entire scale of 3DCNN increases. Therefore, if compression is necessary, we suggest to design large and sparse 3DCNNs rather than tiny and dense ones.

\subsubsection{Data Distribution}\quad

\begin{figure}
\centering
\subfigure[UCF11 (k-fold)]{\includegraphics[width=0.23\textwidth]{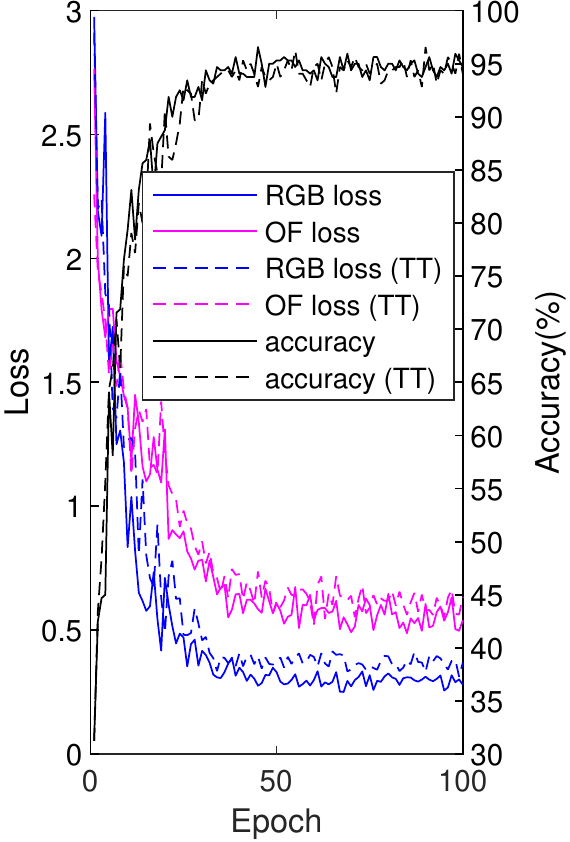}}
\subfigure[UCF101 (split 1)]{\includegraphics[width=0.23\textwidth]{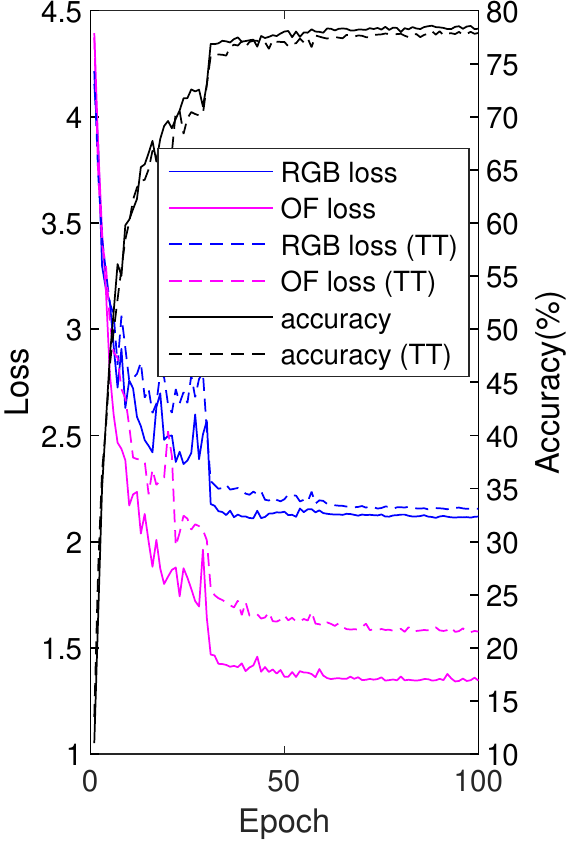}}
\caption{\textbf{Learning curves of two stream 3DCNN on UCF11 and UCF101 datasets.} ``OF'' in the legend means optical flow.}
\label{Fig_learning_curve}
\end{figure}

In macro sense, network redundancy has strong correlation to the data distribution. More challenging dataset may need more redundant network for low loss compression. For example, according to Figure \ref{Fig_learning_curve}, the performances of original and TT networks on UCF11 are still matched, but a little degeneration occurs on UCF101. That is to say, the redundancy of this network is sufficient for UCF11, but not enough for UCF101. Hence, arbitrarily estimating whether a network is redundant or not seems inadvisable. However, researchers can enlarge the size of convolutional kernels and the number of channels brick by brick until their requirements are satisfied.

\subsection{Other Characteristics of TT}

\subsubsection{Regularization}\quad

\begin{figure}
\centering
\includegraphics[width=0.3\textwidth]{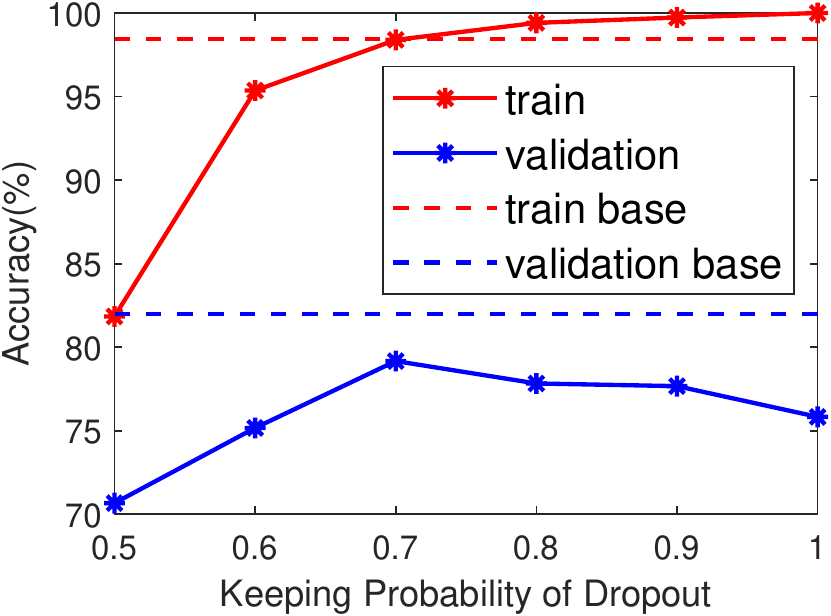}
\caption{\textbf{Testing the regularization of TT decomposition on 3DCNN-VIVA-3.} The dash lines denote the top accuracy of the uncompressed network.}
\label{Fig_regularity}
\end{figure}

It is observed that TT format can bring a certain level of regularization to DNNs. This is the reason why we make the keeping probability of dropout higher in TT networks. Besides, abandoning dropout completely is also inadvisable because even utilizing dropout with 0.9 keeping probability can still avoid over-fitting significantly. For example, we vary the dropout ratio of 3DCNN-VIVA-3 in TT format to observe the regularization effect of TT decomposition in FC layers. The keeping probability of dropout in the uncompressed network is set to 0.5 that illustrates in Figure \ref{Fig_VIVA_3DCNNs}, but the values of keeping factor in the TT network are varied from 0.5 to 1.0. The test result is shown in Figure \ref{Fig_regularity}. We can find that the network presents under-fitting if the keeping probability is less than 0.7, and over-fitting occurs when the keeping probability exceeds 0.7. It is obvious that keeping probability of 0.7 should be the best configuration so that we decide to set this value which can be seen in Figure \ref{Fig_VIVA_3DCNNs}. 

\subsubsection{Latent Degradation}\quad

To be fair, TT decomposition is certain to cause more or less degradation of expressive ability, even the final score may have no degeneration. Thus, accuracy loss will come sooner or later for a DNN with certain scale only if complexity of dataset keeps growing. For instance, according to Figure \ref{Fig_learning_curve}, losses of TT networks are always higher than original networks, and the signs of accuracy loss have appeared from UCF11 to UCF101.

\subsubsection{Computation Complexity}\quad

\begin{figure*}
\centering
\subfigure[Original]{\includegraphics[width=1\textwidth]{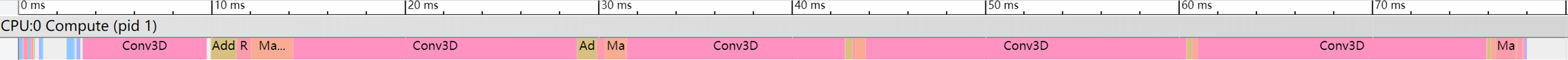}}
\subfigure[TT]{\includegraphics[width=1\textwidth]{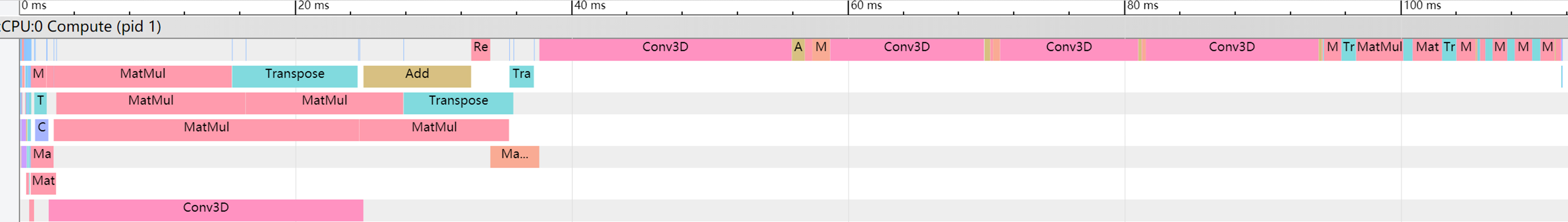}}
\caption{\textbf{Forward executing time of the two stream 3DCNN on UCF11.} These figures are made by the tool ``timeline'' in TensorFlow and run on CPU with 20 batch size. For clarity, ``enisum'' function in t3f is replaced by normal transposition and matrix multiplication, and BN is also disabled.}
\label{Fig_timeline}
\end{figure*}

For a 3D convolutional kernel \(\bm{\mathcal{K}}_{3D} \in \mathbb{R} ^{t \times h \times w \times C \times S}\) with input \(\bm{\mathcal{I}} \in \mathbb{R} ^{T \times H \times W \times C}\), one can easily obtain its computation complexity as \(\mathcal{O}(WHTCS(wht+1))\), and the corresponding computation complexity of TT in \textbf{Algorithm} \ref{Alg_TT_3DCNN} is \(\mathcal{O}(WHTCS(wht+1)+whtCS(r+\sum_{i=1}^{d-1} r^{2}/(cs)^{i}))\), where \(r\) is the maximal TT rank and \(\mathcal{O}(whtCS(r+\sum_{i=1}^{d-1} r^{2}/(cs)^{i}))\) represents the extra amount of calculations which is necessary to recover the TT convolutional kernel to normal format.

Obviously, 3D convolution in TT format has lower computational efficiency than normal format, which is not pointed out clearly by \citet{Garipov_2016_TTCNN}. The reason is that vector matrix multiplication in TT FC layer is completely different from TT convolution process. In fact, vector matrix multiplication is inherently a specific case of mode-(\(N,1\)) contracted product \citep{Lee_2018_TensorNetwork}, thus for a weight matrix in TT format like Equation (\ref{Eq_TTMatrix}), its computational process can be written as
\begin{equation*}
\bm{\mathcal{Y}} = \bm{\mathcal{X}} \times ^{1} \bm{\mathcal{G}}_{1} \times ^{1} \bm{\mathcal{G}}_{2} \times ^{1} \cdots \times ^{1} \bm{\mathcal{G}}_{d}
\end{equation*}
where \(\bm{\mathcal{X}} \in \mathbb{R} ^{m_1 \times m_2 \times \cdots \times m_d}\) and \(\bm{\mathcal{Y}} \in \mathbb{R} ^{n_1 \times n_2 \times \cdots \times n_d}\). The above equation can be calculated from left to right in sequence, so \(\mathcal{O}(dr^{2}\max{\{m,n\}}\max{\{M,N\}})\) should be the computation complexity which avoids recovering weight tensor to original matrix \citep{Novikov_2015_TT}. However, for the computational process in \textbf{Algorithm} \ref{Alg_TT_3DCNN}, it can just be represented like
\begin{equation*}
\bm{\mathcal{O}} = \bm{\mathcal{I}} * (\overline{\bm{\mathcal{G}}}_{0} \times ^{1} \bm{\mathcal{G}}_{1} \times ^{1} \cdots \times ^{1} \bm{\mathcal{G}}_{d})
\end{equation*}
where \(*\) is the convolution operator and there is no associative law between \(*\) and \(\times ^{1}\). Even if we enforce \(\bm{\mathcal{I}}\) to convolute with \(\overline{\bm{\mathcal{G}}}_{0}\) firstly, the subsequent calculations will have no convolutions any more, which certainly can harm feature extraction heavily.

Anyway, the situation is not so bad because of parallel ability of modern hardware. As shown in Figure \ref{Fig_timeline}, we test the executing time of original and TT 3DCNNs on UCF11 by using the tool ``timeline'' in TensorFlow. It can be observed that, in TT network, recovering operations can be parallelly computed in multiple threads, thus the entire executing time of TT 3DCNN has not exceeded that of original network very much. Additionally, in the real practice, recovering process from TT parameters in disk to normal shapes in memory is one-off, i.e., the following calculations will not be influenced by TT at least during forward running.

\subsubsection{Core Significance}\quad

The core significance of using TT decomposition to compress DNNs is that one can easily and directly construct a large network which just consumes tiny storage in the so-called \emph{in situ} training approach \citep{Alibart_2013_ExInSitu} without any delicate design or slow pre-training. That is, TT decomposition affords us a simple and convenient approach to compress DNNs with high compression ratio. Although previous researches \citep{Novikov_2015_TT,Huang_2018_TTCNN,Su_2018_TTCNN,Garipov_2016_TTCNN,Tjandra_2017_TTRNN1,Tjandra_2018_TTRNN2} show that the accuracy loss is hard to avoid, our study demonstrates that 3DCNNs with sufficient redundancy can realize low loss compression based on the TT decomposition.

\section{Conclusions}\label{sec:Con}
This paper introduces a compression method for convolutional kernels in 3DCNNs based on TT decomposition. How to select suitable truncated TT ranks is analyzed and demonstrated in both theory and practice. Our experiments on VIVA challenge, UCF11 and UCF101 datasets verify that 3DCNNs with sufficient redundancy can be compressed in TT format without significant accuracy loss. Moreover, fully utilizing the redundant design for 3DCNNs can result in better performance including higher compression ratio and lower degeneration. Although some latent problems are still to be dealt with, we believe that TT decomposition is a promising approach to compress large scaled 3DCNNs and even other types of DNNs. We would like to point out that currently there are some other compression methods including neural architecture search (NAS) \citep{Zoph_2018_NAS} to automatically optimize the neural network architectures, and the methods focus on directly designing more compact network models \citep{Zhang_2018_ShuffleNet}, and data quantization methods \citep{Wu_2018_WAGE} or network sparsification \citep{Zhu_2018_PruneOrNot} methods to respectively reduce the number of data bits or connections/neurons. The proposed method in this work is perpendicular to these schemes. For future works, the joint-way compression across these techniques also emerges to pursue extreme compression. We shall also aim at extending the proposed method to more comprehensive data sets and exploring its advantages in extensive real world applications.

\section*{Acknowledgement}
This work was partially supported by National Key R\& D Program of China (2018AAA0102600, 2018YFE0200200), and Beijing Academy of Artificial Intelligence (BAAI), Tsinghua University Initiative Scientific Research Program, and a grant from the Institute for Guo Qiang, Tsinghua University.

\end{document}